\pdfoutput=1

\documentclass[11pt]{article}

\usepackage[final]{acl}

\usepackage{times}
\usepackage{latexsym}
\usepackage{amssymb} 

\usepackage[T1]{fontenc}

\usepackage[utf8]{inputenc}
\usepackage[hang,flushmargin]{footmisc}

\usepackage{microtype}
\usepackage{graphicx}
\usepackage{arydshln}
\usepackage{inconsolata}
\usepackage{todonotes}
\usepackage{tabularx}
\usepackage{booktabs} 
\usepackage{multirow}
\usepackage{amsmath}
\usepackage{amssymb}
\usepackage{graphicx}
\usepackage{subcaption}
\usepackage{fge}
\usepackage{colortbl}

\usepackage[framemethod=TikZ]{mdframed} 
\usepackage{listings}
\usepackage{longtable}
\usepackage{tcolorbox}
\usepackage{latexsym}
\usepackage{url}
\usepackage{times}
\usepackage{latexsym}
\usepackage{amsmath}
\usepackage{amssymb} 
\usepackage{url}

\usepackage{pifont} 
\usepackage{url}

\newcommand{\heart}{\ding{170}} 
\newcommand{\spade}{\ding{171}} 
\newcommand{\diamondshape}{\ding{169}} 
\newcommand{\club}{\ding{168}} 

\newcommand{\rparagraph}[1]{\vspace{1.2mm}\noindent\textbf{#1.}}

\newcommand{\rparagraphnodot}[1]{\vspace{1.2mm}\noindent\textbf{#1}}

\definecolor{Gray}{gray}{0.92}
\definecolor{racing-green}{rgb}{0.0, 0.8, 0.6}
\definecolor{awesome-red}{rgb}{1.0, 0.13, 0.32}
\definecolor{LightCyan}{rgb}{0.88,1,1}
\definecolor{darkgreen}{RGB}{0,150,0}

\newcommand{\mydashline}{
  \addlinespace[0.5ex]
  \cdashline{1-10}
  \addlinespace[0.5ex]
}

\newcommand{\ie}{\textit{i}.\textit{e}.,\ }
\newcommand{\eg}{\textit{e}.\textit{g}.,\ }
\newtheorem{definition}{Definition}

\newcolumntype{B}{>{\columncolor{blue!4}}c}
\newcolumntype{d}{>{\columncolor{yellow!4}}c}
\newcolumntype{q}{>{\columncolor{green!4}}c}
\newcolumntype{g}{>{\columncolor{gray!4}}l}

\title{Atomic Calibration of LLMs in Long-Form Generations}

\author{
Caiqi Zhang\textsuperscript{\heart\dag}, Ruihan Yang\textsuperscript{\spade\dag}, Zhisong Zhang\textsuperscript{\diamondshape}\thanks{Corresponding authors.}, \\
\textbf{Xinting Huang}\textsuperscript{\diamondshape}, 
\textbf{Sen Yang}\textsuperscript{\club\dag}, 
\textbf{Dong Yu}\textsuperscript{\diamondshape}, \textbf{Nigel Collier}\textsuperscript{\heart}\footnotemark[1] \\
\textsuperscript{\heart}University of Cambridge,
\textsuperscript{\spade}Fudan University,
\textsuperscript{\diamondshape}Tencent AI Lab, \\
\textsuperscript{\club}The Chinese University of Hong Kong \\
\texttt{\{cz391, nhc30\}@cam.ac.uk, zhisonzhang@tencent.com}
}

\begin{document}
\maketitle

\renewcommand{\thefootnote}{\dag}%
\footnotetext{Work done during the internship at Tencent AI Lab.}

\renewcommand{\thefootnote}{\arabic{footnote}}
\setcounter{footnote}{0}

\begin{abstract}
Large language models (LLMs) often suffer from hallucinations, posing significant challenges for real-world applications. Confidence calibration, as an effective indicator of hallucination, is thus essential to enhance the trustworthiness of LLMs. Prior work mainly focuses on short-form tasks using a single response-level score (macro calibration), which is insufficient for long-form outputs that may contain both accurate and inaccurate claims.
In this work, we systematically  study atomic calibration, which evaluates factuality calibration at a fine-grained level by decomposing long responses into atomic claims. We further categorize existing confidence elicitation methods into discriminative and generative types, and propose two new confidence fusion strategies to improve calibration. Our experiments demonstrate that LLMs exhibit poorer calibration at the atomic level during long-form generation.
More importantly, atomic calibration uncovers insightful patterns regarding the alignment of confidence methods and the changes of confidence throughout generation. This sheds light on future research directions for confidence estimation in long-form generation.
\end{abstract}

\section{Introduction}

\begin{figure}[t]
    \centering
    \includegraphics[width=0.98\columnwidth]{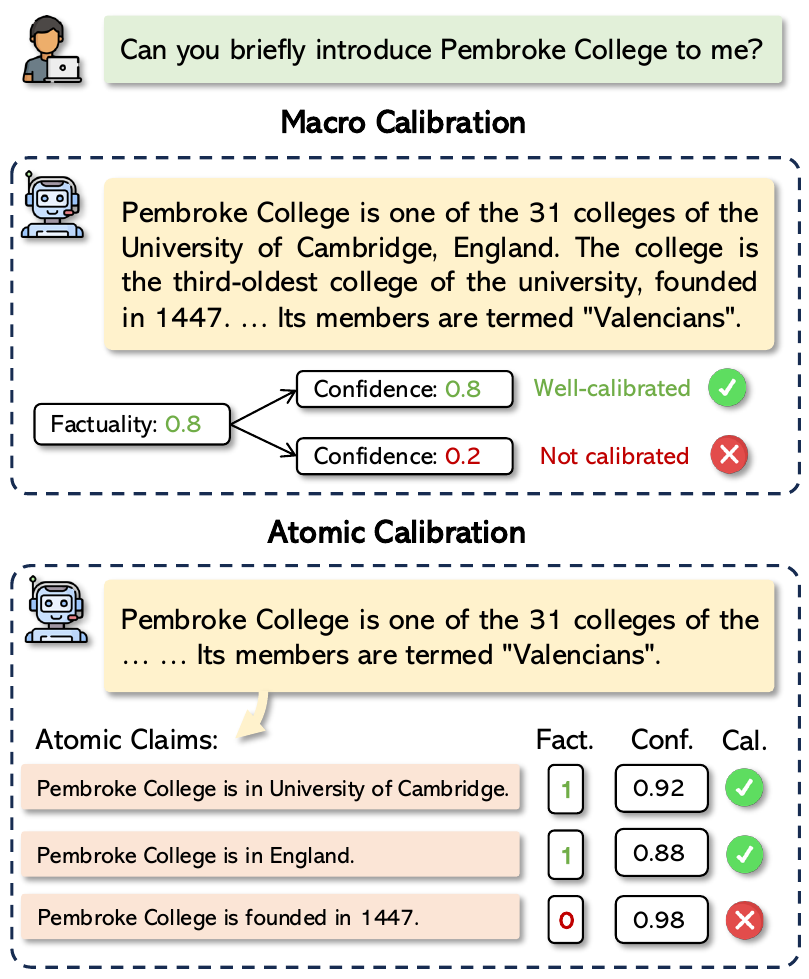}
    \caption{Comparison between traditional macro calibration in response-level and our atomic calibration. The \texttt{Fact.} label is assigned by fact-checking module. We only list three atomic claims for illustration.}
    \label{fig:main}
    \vspace{-2mm}
\end{figure}

While large language models (LLMs) \citep{touvron2023llama, jiang2023mistral, gpt3.5} excel in various tasks, they still struggle with trustworthiness issues.
LLMs often suffer from hallucinations, generating factually inaccurate content and misleading responses \citep{zhang-etal-2024-need, shelmanov-etal-2025-head, he2025supposedly}, which limits their application in high-risk real-world scenarios \citep{hu2023uncertainty}. 
To address this, \textit{confidence calibration} aims to estimate the underlying uncertainty of model predictions and reflect the true likelihood of correctness \citep{guo2017calibration, zhang-etal-2025-roads, zhang2025reinforcement}. 
In this work, we distinguish between calibration as a \textit{property} and as a \textit{process}. We use the term \textit{atomic calibration} to refer to the property of each atomic claim's confidence being well-aligned with its factuality \citep{zhou2025beyond}. In contrast, the extra post-hoc methods used to achieve this alignment (such as temperature scaling and platt scaling) are referred to as \textit{calibration methods} or a \textit{calibrator}, which is not considered in this paper.

A calibrated model is crucial for real-world applications, as it allows us to determine the extent to which we can trust models' predictions \citep{zhu-etal-2023-calibration, mahaut-etal-2024-factual}. Improved calibration enables more reliable confidence estimation, warning users when \textbf{not} to trust the model and \textit{thus mitigating the impact of hallucinations}.

Most existing work on LLM calibration focuses on short-form QA tasks \citep{jiang-etal-2021-know, tian-etal-2023-just, zhu-etal-2023-calibration, ulmer-etal-2024-calibrating}, using datasets like TriviaQA and Natural Questions \citep{joshi-etal-2017-triviaqa}, where answers are typically under 10 words. In contrast, real-world queries often elicit much longer responses \citep{zhang2024luq, yang2024logu, yang-etal-2025-uncle}, spanning hundreds or thousands of words. In such cases, response quality is not simply binary, as answers may mix accurate and inaccurate statements.

Recent work has begun to address calibration in long-form generation \citep{zhang2024luq, huang2024calibrating, liu2023litcab, fadeeva-etal-2024-fact, jiang2024graphbased}. Several approaches estimate a single confidence score for the entire response (\textit{macro calibration}; upper, Figure~\ref{fig:main}), while others assess confidence at the level of atomic claims \citep{liu2023litcab, fadeeva-etal-2024-fact, jiang2024graphbased} (\textit{atomic calibration}; lower, Figure~\ref{fig:main}). However, previous work leaves several key research questions unanswered: Why is it important to evaluate calibration at the atomic-claim level? What factors influence calibration results at this level? What patterns can be observed by analyzing calibration at the atomic-claim level?

In this work, we systematically examine \textbf{atomic calibration}: A long response is decomposed into atomic claims, each containing a single factual statement, and confidence scores are assigned using various elicitation methods. To analyze different confidence elicitation methods, we categorize them into \textbf{discriminative} (intrinsic confidence estimation) and \textbf{generative} (external confidence assessment). 
Our experiments on three long-form QA datasets with seven LLMs reveal that: \textbf{(1)} Models that appear \textit{well-calibrated} at the response level \textit{perform poorly at the atomic level} (Figure \ref{fig:ece_brier_score}, Table \ref{tab:atomic_main}); \textbf{(2)} Leveraging atomic calibration enhances macro calibration (Table \ref{tab:response_main}). These two reasons highlight the need for research on atomic-level calibration to develop better-calibrated models.

We further investigate the characteristics of discriminative and generative confidence. Our analysis yields two main findings: \textbf{(1)} Discriminative and generative methods are \textit{complementary}; combining them improves calibration, while combinations within the same category offer limited gains. \textbf{(2)} \textit{Generative methods} maintain consistent calibration across different model sizes, whereas \textit{discriminative methods} benefit from increased model sizes.
Motivated by finding (1), we propose two novel fusion strategies based on \textit{confidence agreement} to integrate generative and discriminative confidence. Our strategies outperform existing fusion methods.

Our atomic-level analysis (Section~\ref{sec:analysis}) offers deeper insights into \textbf{confidence method alignment} and \textbf{confidence changes during generation}.  Confidence methods within the same category align better, explaining why cross-category fusion is more effective. Interestingly, with discriminative methods, model confidence in atomic facts tends to \textit{decrease} as generation progresses. In contrast, generative methods show the \textit{lowest average confidence in the middle} of the generation process. These results highlight the necessity of fine-grained calibration evaluation for long-form generation, given us insights on model trustworthiness and usability.
\section{Related Work}

\rparagraph{Atomic Claims Generation and Verification} Long-form responses often contain both correct and incorrect statements, which impact the overall factuality assessments. \citet{min-etal-2023-factscore} propose breaking long responses into atomic facts and calculating the precision of these fact pieces to determine the overall factuality score. \citet{wei2024longfact} and \citet{zhao2024wildhallu} extend this paradigm by expanding the dataset to include more domains beyond biographies. \citet{song2024veriscore} design \textsc{VeriScore} for diverse long-form generation tasks that feature both verifiable and unverifiable content. \citet{chiang-lee-2024-merging} introduce \textsc{D-FActScore}, specifically designed for content with ambiguous entities. Decomposing long-form responses into atomic claims and fact-checking them individually has become a widely adopted pipeline.

\rparagraph{Uncertainty and Calibration in Long-form Generations} Existing research on uncertainty estimation and calibration primarily focuses on multiple-choice or short-form questions \citep{zhu-etal-2023-calibration, kuhn2022semantic, lin2023generating, tian-etal-2023-just, ulmer-etal-2024-calibrating}. There is an increasing interest on calibration for long-form generations. \citet{huang2024calibrating} proposed a unified calibration framework for all text generation tasks, comparing distributions of both correctness and the associated confidence of responses. \citet{band2024linguistic} introduced linguistic calibration, where models explicitly express their uncertainty during long-form generation. \citet{zhang2024luq} proposed \textsc{Luq}, an uncertainty estimation method tailored to long-form generation, demonstrating its effectiveness in ensembling different LLMs. 
Another line of work \citep{liu2023litcab, fadeeva-etal-2024-fact, jiang2024graphbased, yuan2024factlevel} decomposes sentences into atomic claims and assigns confidence scores to each claim. However, a unified definition of atomic-level calibration remains lacking. Clarifying this concept and identifying key influencing factors are essential steps toward improving calibration in long-form generation.
\section{Atomic Calibration}
\label{sec:atomic}

For a language model \( \mathcal{M} \), let $ x \sim M(x\mid q)$ denote the response generated by \( \mathcal{M} \) for a query \( q \), \( x \in \mathcal{X} \). Let \( y \in \mathcal{Y}_{t} \) be the corresponding label, representing a quality score ranging from 0 to 1 for a specific task \( t \in T \). Unlike multiple-choice or short-form questions, which mainly assess correctness, tasks in \( T \) cover diverse dimensions such as factuality, coherence, and creativity.

We define a probability prediction function \( f: \mathcal{X} \to \Delta^{\lvert \mathcal{Y}_{t} \rvert} \), where \( \Delta^{\lvert \mathcal{Y}_{t} \rvert} \) denotes the \( \lvert \mathcal{Y}_{t} \rvert \)-dimensional probability simplex. Here, \( f(x)_y \) represents the probability assigned to label \( y \) for a generated output \( x \). 
In this work, we focus on calibrating factuality, as hallucinations are a well-known issue in LLMs \citep{zhang2023siren, huang2023survey}, and the factuality of atomic claims can be assessed objectively. In this setting, \( \mathcal{Y} \) denotes \( \mathcal{Y}_{t} \) for the factuality task \( t \), where \( \mathcal{Y} \subseteq [0, 1] \) reflects the factuality level of a response. 
Following \citet{guo2017calibration}, we define the calibration of each response as follows:

\begin{definition}[Macro Calibration on Factuality]
\setlength{\abovedisplayskip}{10pt}
\setlength{\belowdisplayskip}{-4pt}
A language model \( \mathcal{M} \) that produces generations \( x \sim \mathcal{M}(x \mid q) \) is said to be \textbf{response-level (macro) calibrated} if
\[
\mathbb{P}( y \mid f(x)_{y} = \beta ) = \beta, \quad \forall \beta \in [0,1].
\]
\label{defn:resp_cal}
\end{definition}

In the context of long-form generation, a single response \( x \) may encompass multiple atomic claims. Macro calibration at the response level cannot fully present the fine-grained uncertainty at the atomic level. 
To address this, we decompose the response \( x \) into $N$ atomic claims \( c_i \), represented as \( x = \coprod_{i=1}^{N} c_{i} \). 
Each atomic claim \( c_i \) is assigned a binary label \( y_i \in \mathcal{Y}_i \), where \( \mathcal{Y}_i = \{0, 1\} \), indicating its truthfulness. The overall factuality score for the response \( y \) is computed as \( y = \frac{1}{N} \sum_{i=1}^{N} y_{i} \). Similarly, we define \( f(c_i)_{y_i} \) as the probability of the label \( y_i \) given the atomic claim \( c_i \).
Building on this decomposition, we propose a fine-grained measure of calibration at the atomic level as follows:

\begin{definition}[Atomic Calibration on Factuality]
\setlength{\abovedisplayskip}{5pt}
\setlength{\belowdisplayskip}{-12pt}
A language model \( \mathcal{M} \), which generates a long-form response \( x \) conditioned on the query \( q \), \( x \sim \mathcal{M}(x \mid q) \), is considered \textbf{atomic-level calibrated} if, for each atomic claim \( c_i \) with its corresponding label \( y_i \), the following condition holds:
\[
\mathbb{P}\left( y_{i} \mid f(c_{i})_{y_{i}}=\beta_{i} \right) = \beta_{i}, \quad \forall \beta_{i} \in [0,1].
\]
\label{defn:atomic_cal}
\end{definition}

\paragraph{Remarks:} 
\textbf{(1)} Unlike traditional classification problems where \( f(x)_y \)  is usually represented as a single log probability of the predicted answer, it is much more challenging to measure model confidence in text generation tasks. Different confidence elicitation methods may yield different predictions of the \( f(x)_y \); therefore, how to design proper elicitation methods is a key problem.
\textbf{(2)} Macro calibration is not equivalent to the sum of atomic calibrations, as illustrated by:
\setlength{\abovedisplayskip}{8pt}

{\small
\begin{equation*}
    \begin{aligned}
    & \mathbb{P}( y \mid f(x)_{y} = \beta ) = \beta \\ \not\Rightarrow \hspace{3pt }
    & \frac{1}{N} \sum_{i=1}^{N} \mathbb{P}\left( y_{i} \mid f(c_{i})_{y_{i}}=\beta_{i} \right) = \beta \\ \not\Rightarrow \hspace{3pt }
    & \mathbb{P}\left( y_{i} \mid f(c_{i})_{y_{i}}=\beta_{i} \right) = \beta, \forall i \in \{1, ...., N\}.
    \end{aligned}
\end{equation*}
}

\section{Confidence Elicitation Methods}
\label{sec:elicitation}

In this section, we define two types of confidence elicitation methods: \textbf{generative} and \textbf{discriminative}. We then introduce two novel confidence fusion strategies that considers confidence agreement when combining confidence scores. For the response $x$ to a query $q$, $x$ is broken into atomic claims $C$. Following previous work~\citep{min-etal-2023-factscore, wei2024longfact, zhao2024wildhallu}, each atomic claim contains a single piece of information and must be self-contained. For generative methods, we sample an additional set of responses $K$, and compare them against the original response $x$. For each atomic claim in $C$, we assign it a confidence score.

\subsection{Generative Methods}
Generative methods assume that the consistency between different generation samples provides a reliable estimation of model uncertainty \citep{zhang2024luq, jiang2024graphbased}. Generally, an additional natural language inference (NLI) model is used to calculate the consistency. In particular, we have the following two variations:

\rparagraph{\textsc{Gen-Binary}} The basic assumption is that if a fact is frequently conveyed when sampled multiple times, the model is considered ``confident'' about that fact. For an atomic claim $c_i$ in $C$, we utilize a NLI model $\mathcal{M}_{\mathrm{NLI}}$ to examine whether $c_i$ is supported or not supported by each of the additional samples. Let $K_s$ be the set of samples supporting $c_i$. Then, the confidence in $c_i$ is calculated as
\begin{equation*}
    Conf(c_i, K) = \frac{\lvert K_{s} \rvert}{\lvert K \rvert}.
\end{equation*}

\rparagraph{\textsc{Gen-Multi}} \textsc{Gen-Multi} assumes that the model is more confident in facts that are \textbf{consistently} expressed. Unlike \textsc{Gen-Binary}, it further divides the ``not supported'' ($K_{ns}$) into ``conflict'' ($K_{c}$) if the fact is presented differently in the sample, and ``not mentioned'' ($K_{nm}$) if the fact is not mentioned in the sample. 
We then calculate the confidence by only considering supporting and conflicting samples:
\begin{equation*}
    Conf(c_i, K) = \frac{\lvert K_{s} \rvert}{\lvert K_{s} \rvert + \lvert K_{c} \rvert}.
\end{equation*}

\subsection{Discriminative Methods}
Discriminative methods assess uncertainties by asking the model itself \citep{tian-etal-2023-just, xiong2023llms}. This is motivated by the findings that models tend to perform better on discriminative tasks \citep{saunders2022selfcritiquing}, and thus they may already possess the capability to estimate the confidence of their own outputs in a discriminative manner.

\rparagraph{\textsc{Dis-Single}} Following \citet{kadavath2022language, tian-etal-2023-just}, we directly ask the model whether one single atomic claim is true or false. 
The probability the model assigns to token ``True'' ($P(true)$) in its generation is viewed as the confidence. 
As each atomic claim is judged individually, one advantage of this method is that there is no cross-claim influences when the model makes confidence judgments.

\rparagraph{\textsc{Dis-Context}} In addition to the method where each claim is judged in a self-contained way, we also consider a setting where additional context is provided. Here, the context denotes the passage where the atomic claim is extracted, or the prompt that generates the response. 
The context helps the model to more accurately locate the atomic claim, and thus potentially leads to better confidence elicitation. $P(true)$, given the context, is then used as the confidence score, just as in \textsc{Dis-Single}.

\rparagraph{\textsc{Dis-Rating}} Instead of using $P(true)$, in \textsc{Dis-Rating}, we directly prompt the model to assign a numerical value representing its confidence in the atomic claim \( c_i \). A score of 0 indicates no confidence, while 10 represents maximum confidence. An alternative approach is to use semantic expressions ranging from ``Very Uncertain'' to ``Very Confident''. However, \citet{tian-etal-2023-just} demonstrate LLMs achieve comparable or even better results using numerical values.

\subsection{Confidence Fusion Strategies}

Combining confidence scores has proven effective for calibration \citep{huang2024calibrating, rivera2024combining}, but existing methods  typically only use a single fixed weight, $\alpha$, to combine the scores, ignoring the \textbf{confidence disagreement}. For instance, \citet{rivera2024combining} computes the weighted average (\textbf{\texttt{WAvg}}) for confidence scores $A$ and $B$:
$
C = A \cdot \alpha + B \cdot (1 - \alpha)
$.
This approach \textit{does not account for the agreement between the two scores}. For example, when $\alpha = 0.5$, confidences of 0 and 1 are treated the same as confidences of 0.4 and 0.6, although the former may indicate higher uncertainty due to a larger disagreement. To address this, we propose two simple but effective methods that consider confidence disagreement $d = B - A$.

\rparagraphnodot{\texttt{AdjustedAlpha}} adjusts the weight $\alpha$ based on the confidence difference:
\vspace{-1mm}
\[
\alpha' = \alpha + \gamma_{\text{a}} \cdot d,
\]
where $\gamma_{\text{a}}$ is a small constant (\eg 0.1), followed by $C' = A \cdot \alpha' + B \cdot (1 - \alpha')$.

\rparagraphnodot{\texttt{DampedFusion}} applies a damping factor based on the agreement:
\vspace{-1mm}
\[
\gamma(d) = 1 - k \cdot |d|,
\]
where $k$ is a small constant (\eg 0.02) that controls the damping sensitivity, followed by
$C' = C \cdot \gamma(d)$. For baselines, we also include: \textbf{\texttt{MinConf}}, which selects the minimum confidence; \textbf{\texttt{HMean}}, which calculates the harmonic mean; and \textbf{\texttt{ProdConf}}, which multiplies the confidences.

\begin{table*}[ht!]
\centering
\setlength\tabcolsep{10pt}
\scalebox{0.92}{
\footnotesize
\renewcommand{\arraystretch}{0.85}  
\begin{tabular}{lBBBdddqqq}
\toprule
 & \multicolumn{3}{c}{\textbf{Bios}} & \multicolumn{3}{c}{\textbf{LongFact}} & \multicolumn{3}{c}{\textbf{WildHallu}} \\
\cmidrule(lr){2-4}  \cmidrule(lr){5-7} \cmidrule(lr){8-10}
\multicolumn{1}{c}{} & \multicolumn{1}{c}{ECE $\downarrow$} & \multicolumn{1}{c}{BS $\downarrow$} & \multicolumn{1}{c}{AUROC $\uparrow$} & \multicolumn{1}{c}{ECE $\downarrow$} & \multicolumn{1}{c}{BS $\downarrow$} & \multicolumn{1}{c}{AUROC $\uparrow$} & \multicolumn{1}{c}{ECE $\downarrow$} & \multicolumn{1}{c}{BS $\downarrow$} & \multicolumn{1}{c}{AUROC $\uparrow$} \\
\midrule
\rowcolor[gray]{1}
\multicolumn{10}{l}{\textbf{Llama3-8B-Instruct}} \\
\midrule
\textsc{Dis-Context} & 35.5 & 35.8 & 74.5 & 11.9 & 13.6 & 74.4 & 12.5 & 16.5 & \textbf{83.5} \\
\textsc{Dis-Rating} & 26.8 & 29.0 & 71.1 & \textbf{3.5} & 12.0 & 66.9 & \textbf{5.3} & \textbf{15.2} & 79.8 \\
\textsc{Dis-Single} & 32.6 & 33.9 & 74.5 & 14.3 & 15.2 & 69.8 & 19.2 & 20.9 & 79.3 \\
\textsc{Gen-Binary} & \textbf{10.0} & \textbf{17.8} & \textbf{83.1} & 8.5 & \textbf{11.4} & \textbf{77.3} & 11.1 & \textbf{15.2} & 82.0 \\
\textsc{Gen-Multi} & 37.4 & 37.3 & 64.2 & 12.6 & 13.1 & 58.5 & 21.9 & 22.1 & 65.4 \\
\midrule
\rowcolor[gray]{1}
\multicolumn{10}{l}{\textbf{Mistral-7B-Instruct}} \\
\midrule
\textsc{Dis-Context} & 24.8 & 26.0 & 77.5 & 15.7 & 16.1 & 75.3 & 20.6 & 21.7 & 79.8 \\
\textsc{Dis-Rating} & 44.5 & 42.5 & 65.0 & 10.0 & 14.2 & 67.9 & 19.7 & 23.9 & 68.1 \\
\textsc{Dis-Single} & 30.2 & 30.7 & 75.2 & 20.4 & 20.5 & 66.6 & 24.0 & 24.6 & 75.1 \\
\textsc{Gen-Binary} & \textbf{13.7} & \textbf{19.0} & \textbf{81.9} & \textbf{8.4} & \textbf{11.5} & \textbf{80.1} & \textbf{12.7} & \textbf{17.0} & \textbf{81.3} \\
\textsc{Gen-Multi} & 42.2 & 41.8 & 65.0 & 13.4 & 13.9 & 61.7 & 26.6 & 26.4 & 64.2 \\
\midrule
\rowcolor[gray]{1}
\multicolumn{10}{l}{\textbf{Qwen2-7B-Instruct}} \\
\midrule
\textsc{Dis-Context} & 26.5 & 28.3 & 75.5 & 13.9 & 14.8 & 77.9 & 17.2 & 19.4 & 81.2 \\
\textsc{Dis-Rating} & 41.5 & 39.7 & 64.2 & \textbf{3.5} & 11.7 & 62.6 & \textbf{8.2} & 18.1 & 70.4 \\
\textsc{Dis-Single} & 29.3 & 30.4 & 75.5 & 16.1 & 16.8 & 74.7 & 18.7 & 20.3 & 80.1 \\
\textsc{Gen-Binary} & \textbf{10.9} & \textbf{16.7} & \textbf{83.8} & 6.3 & \textbf{9.9} & \textbf{81.9} & 9.5 & \textbf{14.0} & \textbf{82.5} \\
\textsc{Gen-Multi} & 41.7 & 41.1 & 65.6 & 11.6 & 12.1 & 62.8 & 21.0 & 21.0 & 64.4 \\
\bottomrule
\end{tabular}
}
\caption{Atomic Calibration Results. All the numbers are in percentages.}
\vspace{-4mm}
\label{tab:atomic_main}
\end{table*}

\section{Experiments and Results}
\label{sec:expriment}

\subsection{Experiment Setup}

\rparagraph{Models} We utilize seven LLMs from three model families with varying sizes: Llama3 Instruct (8B and 70B) \citep{llama3modelcard}, Mistral Instruct (7B and 8x7B) \citep{jiang2023mistral}, and Qwen2 Instruct (7B, 52B-A14B, and 72B) \citep{yang2024qwen2technicalreport}.

\rparagraph{Datasets} We use three datasets for long-form QA: \textit{Bios} \citep{min-etal-2023-factscore}, which contains 500 individuals from Wikipedia with varying levels of popularity, for which models are tasked to generate biographies; \textit{LongFact} \citep{wei2024longfact} extends \textit{Bios} and includes 1,140 questions covering 38 manually-selected topics; \textit{WildHallu} \citep{zhao2024wildhallu} includes 7,917 entities derived from one million user-chatbot interactions in real-world settings. 

\rparagraph{Atomic Facts Generation and Verification} For all three datasets, we apply a \textsc{FActScore}-based \citep{min-etal-2023-factscore} factuality assessment approach. We first use GPT-4o to decompose the entire response into atomic facts. These atomic facts are then verified using GPT-4o, cross-referenced with evidence from Wikipedia and Google Search. The detailed prompts for generating atomic facts are provided in Appendix \ref{app:prompts}.

\rparagraph{Confidence Elicitation} We use P(true) \citep{kadavath2022language}, Self-Rating \citep{tian-etal-2023-just}, Semantic Entropy (SE) \citep{kuhn2022semantic}, and Sum of Eigenvalues (EigV) \citep{lin2023generating} as the baseline confidence elicitation methods. They are all calculated in response-level. For \textsc{Gen-Binary}, 
we apply the Llama-3-8B-Instruct for better NLI performance. For \texttt{WAvg, AdjustedAlpha}, and \texttt{DampedConf}, we use a separate validation set for hyper-parameter tuning.

\rparagraph{Metrics} We use Expected Calibration Error (ECE) \citep{naeini2015obtaining} and Brier Score (BS) \citep{brier1950verification} as the primary metrics. These metrics are applicable to both atomic and macro calibration (see details in Appendix \ref{app:atomic_metrics}), enabling a direct comparison between them. Additionally, we include AUROC to evaluate atomic calibration and Spearman Correlation for a more instance-specific assessment in macro calibration.

\begin{figure}[th!]
    \centering
    \includegraphics[width=\columnwidth]{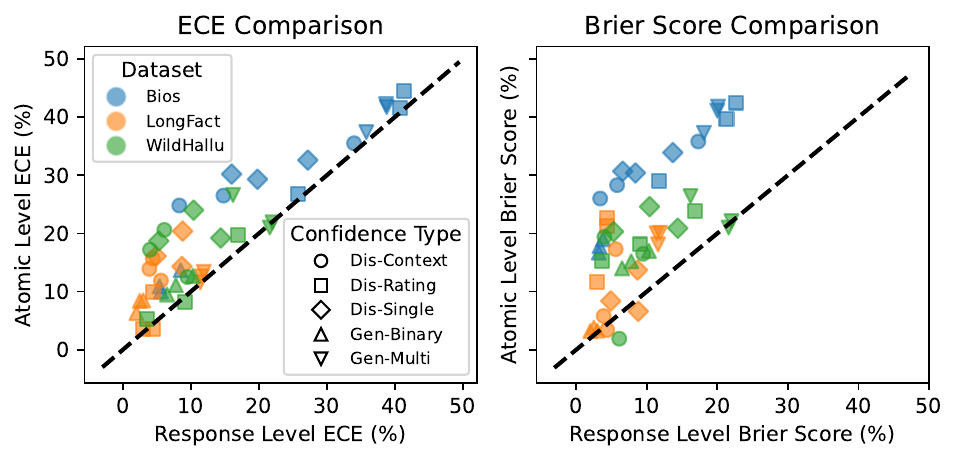}
    \caption{Comparison of atomic level and response-level calibration for ECE and Brier Score. Atomic-level performance is generally \textbf{worse} than response-level performance, with data points consistently lying \textbf{above} the identity line.}
    \label{fig:ece_brier_score}
\end{figure}

\begin{table*}[ht!]
\centering
\renewcommand{\arraystretch}{0.82}  
\scalebox{0.82}{
\footnotesize
\setlength{\tabcolsep}{15pt} 
\begin{tabular}{lBBBdddqqq}
\toprule
 & \multicolumn{3}{c}{\textbf{Bios}} & \multicolumn{3}{c}{\textbf{LongFact}} & \multicolumn{3}{c}{\textbf{WildHallu}} \\
\cmidrule(lr){2-4}  \cmidrule(lr){5-7} \cmidrule(lr){8-10}
\multicolumn{1}{c}{} & \multicolumn{1}{c}{ECE $\downarrow$} & \multicolumn{1}{c}{BS $\downarrow$} & \multicolumn{1}{c}{SC $\uparrow$} & \multicolumn{1}{c}{ECE $\downarrow$} & \multicolumn{1}{c}{BS $\downarrow$} & \multicolumn{1}{c}{SC $\uparrow$} & \multicolumn{1}{c}{ECE $\downarrow$} & \multicolumn{1}{c}{BS $\downarrow$} & \multicolumn{1}{c}{SC $\uparrow$} \\
\midrule
\rowcolor[gray]{1}
\multicolumn{10}{l}{\textbf{Llama3-8B-Instruct}} \\
\midrule

P(true) & 45.1 & 25.9 & 30.2 & 16.3 & 4.8 & 18.9 & 25.7 & 13.5 & 40.5 \\
Self-Rating & 38.7 & 23.4 & 40.5 & 14.1 & 4.2 & 21.5 & 18.6 & 12.9 & 50.2 \\
SE & 37.4 & 21.8 & 42.1 & 13.5 & 3.4 & 23.0 & 17.8 & 11.7 & 52.0 \\
EigV & 36.8 & 21.2 & 43.0 & 13.0 & 3.2 & 23.8 & 17.2 & 11.3 & 53.0 \\
\mydashline

\textsc{Dis-Context} & 34.0 & 17.3 & 55.4 & 5.6 & 1.9 & 29.7 & 9.5 & 4.8 & 65.9 \\
\textsc{Dis-Rating} & 25.7 & 11.7 & 73.8 & \textbf{2.9} & 1.6 & 34.1 & \textbf{3.6} & \textbf{3.5} & \textbf{71.7} \\
\textsc{Dis-Single} & 27.2 & 13.7 & 58.0 & 8.7 & 2.6 & 20.9 & 14.4 & 7.3 & 55.9 \\
\textsc{Gen-Binary} & \textbf{5.6} & \textbf{3.3} & \textbf{79.8} & 3.0 & \textbf{1.1} & \textbf{52.7} & 7.8 & 4.6 & 70.0 \\
\textsc{Gen-Multi} & 35.8 & 18.1 & 71.4 & 11.6 & 2.7 & 37.5 & 22.0 & 10.5 & 62.6 \\
\midrule
\rowcolor[gray]{1}
\multicolumn{10}{l}{\textbf{Mistral-7B-Instruct}} \\
\midrule
P(true) & 44.5 & 27.1 & 32.8 & 16.7 & 7.4 & 22.0 & 24.3 & 19.8 & 41.2 \\
Self-Rating & 37.1 & 26.4 & 42.3 & 14.5 & 6.5 & 26.1 & 18.1 & 14.5 & 52.0 \\
SE & 36.5 & 23.9 & 44.1 & 13.8 & 3.7 & 28.0 & 17.4 & 14.3 & 53.4 \\
EigV & 35.9 & 23.3 & 45.3 & 13.3 & 3.5 & 36.5 & 16.9 & 13.9 & 54.4 \\
\mydashline

\textsc{Dis-Context} & \textbf{8.3} & \textbf{3.4} & \textbf{79.7} & 4.1 & 1.4 & 47.9 & \textbf{6.1} & \textbf{4.3} & 72.3 \\
\textsc{Dis-Rating} & 41.4 & 22.7 & 55.0 & 4.4 & 1.7 & 40.8 & 16.9 & 9.8 & 60.4 \\
\textsc{Dis-Single} & 16.0 & 6.6 & 70.3 & 8.8 & 3.1 & 32.8 & 10.4 & 6.5 & 65.3 \\
\textsc{Gen-Binary} & 8.5 & 3.8 & 74.9 & \textbf{2.5} & \textbf{1.0} & \textbf{64.1} & 10.3 & 5.0 & \textbf{73.9} \\
\textsc{Gen-Multi} & 38.7 & 20.1 & 60.7 & 11.9 & 2.8 & 49.6 & 26.2 & 13.4 & 65.6 \\
\midrule
\rowcolor[gray]{1}
\multicolumn{10}{l}{\textbf{Qwen2-7B-Instruct}} \\
\midrule

P(true)        & 45.0 & 27.9 & 33.5 & 11.2 & 5.6 & 28.3 & 12.7 & 15.6 & 35.4 \\
Self-Rating    & 24.3 & 25.1 & 48.2 & 6.9  & 4.9 & 36.7 & 9.8  & 14.7 & 48.0 \\
SE             & 22.9 & 23.1 & 49.8 & 6.5  & 3.7 & 38.9 & 8.9  & 13.9 & 49.2 \\
EigV           & 22.4 & 22.5 & 50.7 & 6.2  & 3.5 & 39.8 & 8.5  & 13.5 & 50.2 \\
\mydashline

\textsc{Dis-Context} & 14.8 & 5.8 & 66.5 & 3.9 & 1.7 & 40.6 & \textbf{4.0} & 3.5 & 66.8 \\
\textsc{Dis-Rating} & 40.7 & 21.3 & 63.0 & 4.4 & 1.9 & 29.9 & 9.1 & 6.3 & 54.0 \\
\textsc{Dis-Single} & 19.8 & 8.4 & 52.8 & 4.9 & 2.4 & 30.9 & 5.3 & 4.8 & 60.4 \\
\textsc{Gen-Binary} & \textbf{5.4} & \textbf{3.2} & \textbf{72.4} & \textbf{2.0} & \textbf{0.9} & \textbf{67.6} & 6.5 & \textbf{3.2} & \textbf{72.2} \\
\textsc{Gen-Multi} & 38.8 & 20.0 & 43.1 & 11.4 & 2.6 & 52.1 & 21.6 & 9.5 & 63.2 \\
\bottomrule
\end{tabular}
}
\caption{Macro Calibration Results. All the numbers are in percentages.}
\label{tab:response_main}
\vspace{-2mm}
\end{table*}

\subsection{Results}

\rparagraph{Overall, the tested LLMs are not well-calibrated at the atomic fact level} Table \ref{tab:atomic_main} lists our main atomic calibration results. Although there is no universally accepted threshold for low ECE, a well-calibrated model typically achieves an ECE close to 1\%, as shown in \citep{guo2017calibration} and \citep{zhu-etal-2023-calibration}. However, even with the most robust method, \textsc{Gen-Binary}, the ECE scores remain around 10\%, indicating a significant calibration gap. Among the models, Qwen2-7B-Instruct demonstrates slightly better calibration compared to the other two. 

\paragraph{Models that appear well-calibrated at the response level still perform poorly at the atomic level} Figure \ref{fig:ece_brier_score} compares atomic and response-level scores for ECE and Brier Score across different datasets and confidence types. The data points consistently lie above the identity line, indicating that \textit{atomic-level errors are higher than response-level errors}. This suggests that atomic calibration is crucial for fine-grained evaluation.

\rparagraph{Atomic calibration can enhance macro calibration} Table \ref{tab:response_main} shows the main results of response-level calibration. For the five atomic-level methods, we calculate the average confidence of the facts in a response to obtain the response-level confidence. The results indicate that atomic calibration leads to better overall results compared to the baseline methods, highlighting the helpfulness of more fine-grained calibration analysis.

\rparagraph{The confidence fusion method considering confidence agreement outperforms other methods} Table \ref{tab:fusion_paper} presents the results of various confidence fusion strategies at the atomic level (more results in Appendix \ref{app:fusions}). The best performance is consistently achieved by \texttt{AdjustedAlpha} and \texttt{DampedFusion}. Notably, we observe that combining methods of the same confidence type (e.g., \textsc{Dis-Rating} with \textsc{Dis-Context}) \textit{does not lead to improved calibration}. A case study demonstrating the effectiveness of confidence fusion is shown in Figure \ref{fig:case_study}.

\rparagraph{Larger model size does not necessarily result in better calibration} Table \ref{tab:atomic_sclaing} 
compares the calibration levels of models with different sizes. Our two key findings are: \textbf{(1)} With generative methods, there is little difference in calibration between larger and smaller models; \textbf{(2)} With discriminative methods, \textit{larger models generally provide better calibration}. We hypothesize that this is because discriminative methods require models to self-assess the confidence of their own outputs, and larger models typically possess stronger discriminative abilities \citep{saunders2022selfcritiquing}.

\begin{table*}[h!]
\centering
\setlength\tabcolsep{9pt}
\scalebox{0.9}{
\footnotesize
\begin{tabular}{lBBBdddqqq}
\toprule
 & \multicolumn{3}{c}{\textbf{Bios}} & \multicolumn{3}{c}{\textbf{LongFact}} & \multicolumn{3}{c}{\textbf{WildHallu}} \\
\cmidrule(lr){2-4}  \cmidrule(lr){5-7} \cmidrule(lr){8-10}
\multicolumn{1}{c}{} & \multicolumn{1}{c}{ECE $\downarrow$} & \multicolumn{1}{c}{BS $\downarrow$} & \multicolumn{1}{c}{AUROC $\uparrow$} & \multicolumn{1}{c}{ECE $\downarrow$} & \multicolumn{1}{c}{BS $\downarrow$} & \multicolumn{1}{c}{AUROC $\uparrow$} & \multicolumn{1}{c}{ECE $\downarrow$} & \multicolumn{1}{c}{BS $\downarrow$} & \multicolumn{1}{c}{AUROC $\uparrow$} \\
\midrule
\multicolumn{10}{l}{\textbf{\textsc{Gen-Binary}}} \\
\midrule
Llama3-8B-Instruct & 10.0 & 17.8 & 83.1 & 8.5 & 11.4 & 77.3 & 11.1 & 15.2 & 82.0 \\
Llama3-70B-Instruct & 10.0 & 16.5 & 82.5 & 8.3 & 9.3 & 73.7 & 9.5 & 12.3 & 78.3 \\
\mydashline
Mistral-7B-Instruct & 13.7 & 19.0 & 81.9 & 8.4 & 11.5 & 80.1 & 12.7 & 17.0 & 81.3 \\
Mistral-8x7B-Instruct & 12.3 & 18.5 & 79.8 & 7.8 & 9.0 & 76.3 & 9.8 & 13.4 & 77.8 \\
\mydashline
Qwen2-7B-Instruct & 10.9 & 16.7 & 83.8 & 6.3 & 9.9 & 81.9 & 9.5 & 14.0 & 82.5 \\
Qwen2-57B-Instruct & 10.5 & 18.1 & 82.3 & 7.8 & 10.0 & 78.3 & 9.2 & 13.6 & 81.7 \\
Qwen2-72B-Instruct & 11.2 & 16.6 & 83.4 & 7.6 & 8.3 & 76.6 & 8.6 & 11.9 & 77.7 \\
\midrule
\multicolumn{10}{l}{\textbf{\textsc{Dis-Rating}}} \\
\midrule
Llama3-8B-Instruct & 26.8 & 29.0 & 71.1 & 3.5 & 12.0 & 66.9 & 5.3 & 15.2 & 79.8 \\
Llama3-70B-Instruct & 10.6 & 19.3 & 73.2 & 4.2 & 8.0 & 74.2 & 4.3 & 11.5 & 81.2 \\
\mydashline
Mistral-7B-Instruct & 44.5 & 42.5 & 65.0 & 10.0 & 14.2 & 67.9 & 19.7 & 23.9 & 68.1 \\
Mistral-8x7B-Instruct & 15.3 & 22.6 & 70.8 & 5.3 & 8.6 & 72.6 & 7.6 & 14.7 & 72.9 \\
\mydashline
Qwen2-7B-Instruct & 41.5 & 39.7 & 64.2 & 3.5 & 11.7 & 62.6 & 8.2 & 18.1 & 70.4 \\
Qwen2-57B-Instruct & 23.2 & 27.0 & 69.3 & 2.2 & 9.8 & 71.3 & 5.2 & 15.2 & 77.2 \\
Qwen2-72B-Instruct & 11.4 & 21.0 & 71.6 & 6.1 & 7.7 & 77.1 & 4.0 & 11.7 & 79.2 \\
\bottomrule

\end{tabular}
}
\caption{Atomic calibration results with different model sizes. All the numbers are in percentages.}
\label{tab:atomic_sclaing}
\end{table*}

\begin{table*}[t!]
\centering
\setlength\tabcolsep{10pt}
\scalebox{0.9}{
\footnotesize \centering
\begin{tabular}{lBBBdddqqq}
\toprule
 & \multicolumn{3}{c}{\textbf{Bios}} & \multicolumn{3}{c}{\textbf{LongFact}} & \multicolumn{3}{c}{\textbf{WildHallu}} \\
\cmidrule(lr){2-4}  \cmidrule(lr){5-7} \cmidrule(lr){8-10}
\multicolumn{1}{c}{} & \multicolumn{1}{c}{ECE $\downarrow$} & \multicolumn{1}{c}{BS $\downarrow$} & \multicolumn{1}{c}{AUROC $\uparrow$} & \multicolumn{1}{c}{ECE $\downarrow$} & \multicolumn{1}{c}{BS $\downarrow$} & \multicolumn{1}{c}{AUROC $\uparrow$} & \multicolumn{1}{c}{ECE $\downarrow$} & \multicolumn{1}{c}{BS $\downarrow$} & \multicolumn{1}{c}{AUROC $\uparrow$} \\
\midrule
\textsc{Gen-Binary} & 10.0 & 17.8 & 83.1 & 8.5 & 11.4 & 77.3 & 11.1 & 15.2 & 82.0 \\
\textsc{Dis-Rating} & 26.8 & 29.0 & 71.1 & 3.5 & 12.0 & 66.9 & 5.3 & 15.2 & 79.8 \\
\textsc{Dis-Context} & 35.5 & 35.8 & 74.5 & 11.9 & 13.6 & 74.4 & 12.5 & 16.5 & 83.5 \\

\addlinespace[0.5ex]
\cdashline{1-10}
\addlinespace[0.5ex]

\texttt{MinConf} & 6.2 & 17.1 & 83.2 & 10.7 & 12.2 & 77.4 & 9.0 & 13.9 & 85.8 \\
\texttt{HMean} & 9.8 & 17.4 & 84.0 & 4.1 & 11.0 & 79.6 & 5.6 & 13.3 & 87.0 \\
\texttt{ProdConf} & 7.4 & 16.7 & 84.1 & 13.4 & 12.8 & 79.6 & 11.5 & 14.1 & 87.0 \\
\texttt{WAvg} & 10.9 & 17.4 & 84.4 &\textbf{ 3.3} & 10.3 & 79.9 & 5.1 & 13.0 & 87.0 \\

\addlinespace[0.5ex]
\cdashline{1-10}
\addlinespace[0.5ex]

\texttt{AdjustedAlpha} & \textbf{4.1} & 15.8 & \textbf{85.2} & 3.4 & 10.2 & \textbf{80.4} & \textbf{4.3} & 12.6 & \textbf{88.3} \\
\texttt{DampedFusion} & 5.0 & \textbf{15.6} & 84.7 & 3.5 & \textbf{9.8} & 80.0 & 4.8 & \textbf{12.4} & 87.9 \\
\bottomrule
\end{tabular}
}
\caption{Atomic calibration results of different confidence fusion strategies for Llama3-8B-Instruct. The fusion results are based on \textsc{Gen-Binary} and \textsc{Dis-Rating}.}
\label{tab:fusion_paper}
\vspace{-2mm}
\end{table*}

\section{Discussion} \label{sec:analysis}
\subsection{Confidence Methods Alignment} \label{sec:alignment}

\begin{figure}[h!]
    \centering
    \includegraphics[width=1\columnwidth]{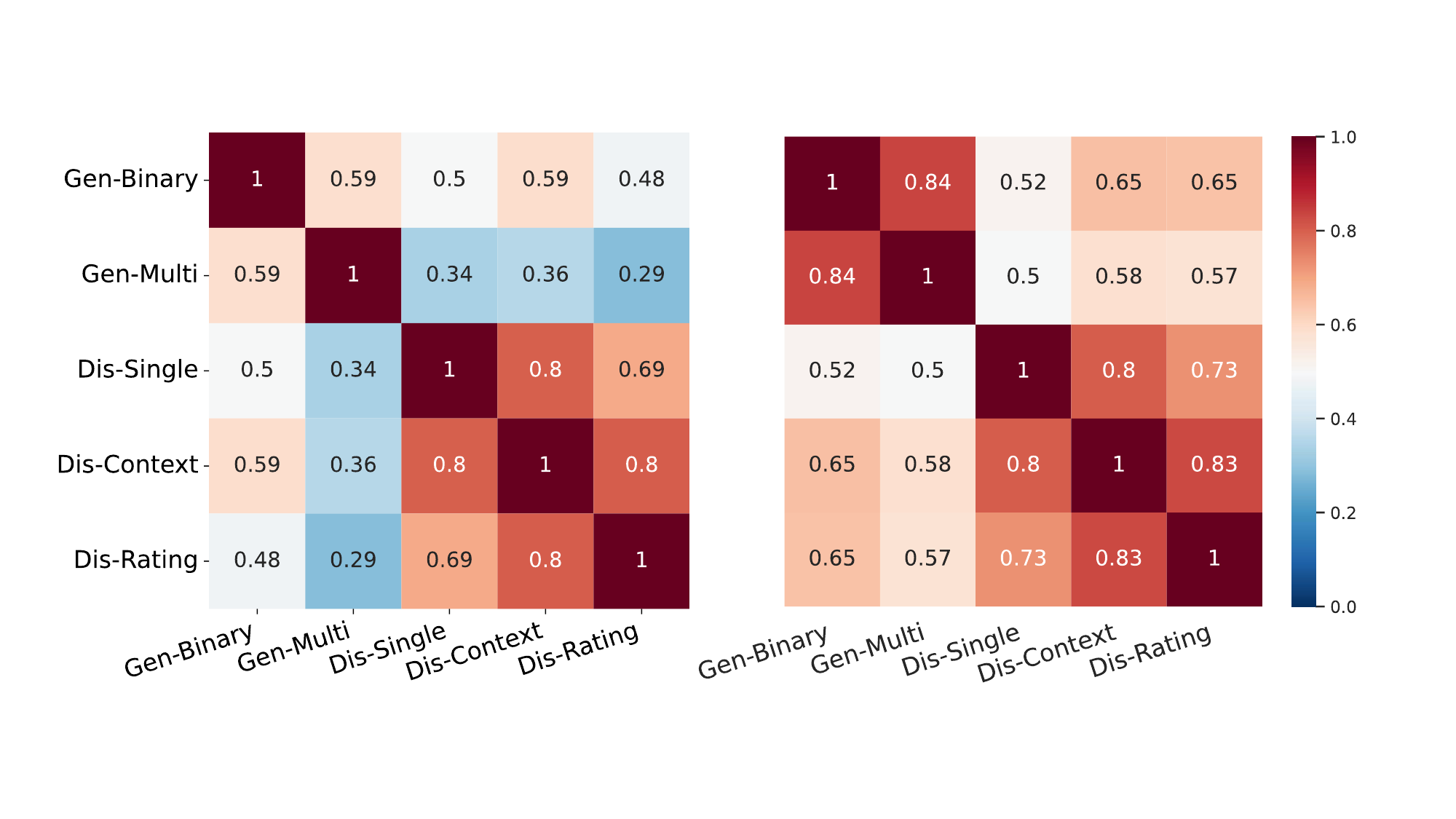}
    \vspace{-2mm}
    \caption{Heatmaps of Spearman Correlation between different confidences in Llama3-8B-Instruct on \textit{WildHallu}. Warmer colors indicate higher correlations. Atomic level: left; response level: right.}
    \label{fig:alignment}
    \vspace{-2mm}
\end{figure}

To further explore the reasons behind the improvements provided by confidence fusion, we show the correlation between different confidence elicitation methods in Figure \ref{fig:alignment} (using \textit{WildHallu} as the study case and more results are in Appendix \ref{app:conf_align}). Our findings are summarized as follows:

\rparagraph{Confidence methods within the same type are better aligned} In Figure \ref{fig:alignment}, warmer colors indicate higher Spearman Correlation scores. Confidence elicitaton methods of the same type (top left for generative and bottom right for discriminative) show stronger correlations compared to those across different types. This helps to explain why cross-category fusion strategies are effective, since these two types \textit{capture different aspects of uncertainty} and are complementary to each other.

\rparagraph{The alignment is stronger at the response level than at the atomic level} When comparing atomic and macro calibration, we observe that the alignment is stronger for the latter. In atomic calibration, several methods display weak correlations (indicated in blue), while the correlations are generally higher at response level (indicated in red). Similarly, methods from different types show more disagreement than those of the same type.
This highlights the need for future research on the discrepancies between generative and discriminative confidence elicitation methods, as well as how to better unify these approaches.


\begin{figure*}[t!]
    \centering
    \includegraphics[width=1\textwidth]{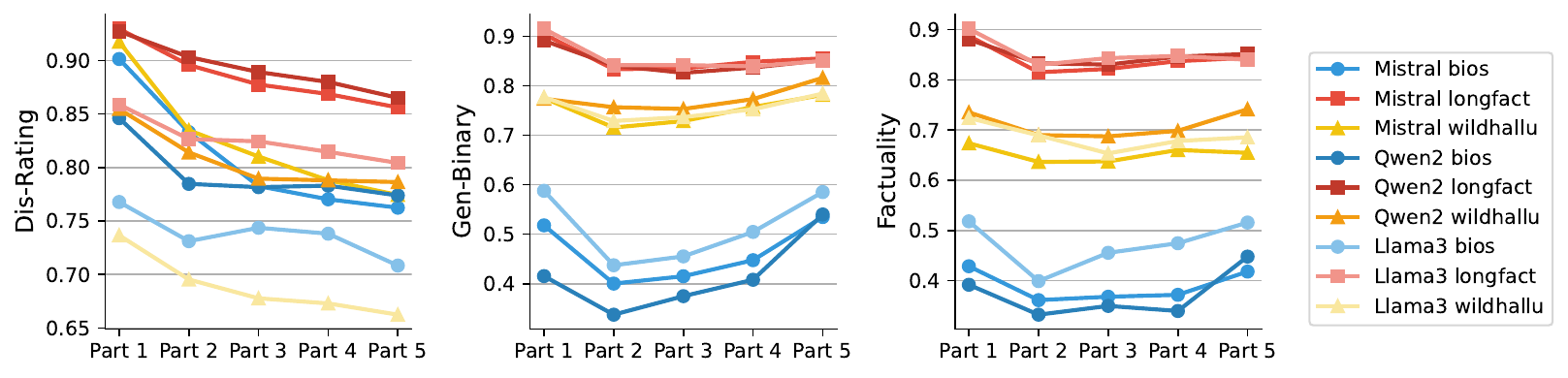}
    \caption{Average confidence scores across different parts of long-form responses. For discriminative methods, confidence decreases as the generation progresses, while generative methods show the lowest confidence in the middle sections. }
    \label{fig:positions}
\end{figure*}

\subsection{Confidence Across Different Positions}
As each long-form response contains multiple atomic facts, we analyze how confidence and factuality scores evolve during the generation process. Specifically, we divide all atomic facts $C$ into five equal parts along the generation process. Part 1 represents the beginning of the generation, and part 5 corresponds to the end. We calculate the average confidence score for each part of the responses and present the results in Figure \ref{fig:positions}.

\rparagraph{With discriminative methods, models exhibit decreasing confidence in atomic facts as the generation progresses} We observe similar trends across all discriminative methods.
This contrasts with previous findings, which used logits as a measure of confidence and found that models tend to become more confident during long generation sequences \citep{zhang-etal-2023-enhancing-uncertainty}. Our results show that discriminative methods indicate lower confidence in the model's output toward the latter parts of the generation.

\paragraph{With generative methods, the model shows the lowest average confidence in the middle part of the generation.} We hypothesize that this is because the tested models tend to provide general introductions and conclusions at the beginning and the end of the generation. During consistency checking, these statements are frequently cross-referenced, leading to higher confidence. For example, in \textit{Bios}, statements like ``\texttt{[a person]} is famous'' or ``\texttt{[a person]} made a significant impact in his field'' are often repeated across samples. On the contrary, in the middle parts where the models address more specific facts about individuals' lives, careers and achievements, they tend to cover different aspects and details.

\subsection{The Utilities of Atomic Calibration}
While the primary goal of atomic calibration is to provide fine-grained calibration evaluation for models, we also explore its utilities in several downstream tasks, including: (1) \textbf{Selective Question Answering} \citep{kamath-etal-2020-selective, cole-etal-2023-selectively, yang2023uncertaintyaware}, which involves setting a confidence threshold to selectively reject low-confidence answers, ensuring that only high-confidence responses are retained; (2) \textbf{LLM-Ensemble} \citep{zhang2024luq}, which leverages multiple models to generate responses to the same question, selecting the answer with the highest confidence, thereby combining the strengths of each model; and (3) \textbf{Atomic Claims Reunion} \citep{thirukovalluru-etal-2024-atomic, jiang2024graphbased}, which involves sampling multiple responses, breaking then into atomic claims, evaluating their confidence, and reassembling only high-confidence claims to produce a more reliable final answer. Among these applications, we observe consistent improvements in factuality with atomic-level examination. Detailed experimental settings and results can be found in Appendix \ref{app:applications}.

It is important to note that, unlike previous work on Selective Question Answering \citep{huang2024calibrating} and LLM-Ensemble \citep{zhang2024luq} for long-form generation, which mainly rely on atomic-level confidence estimation to \textit{enhance the overall quality of responses} (with responses either being entirely accepted or rejected), \textit{Atomic Claims Reunion} \textbf{does not} require an overall response-level score. Instead, it relies entirely on the confidence of atomic claims to select and combine the most accurate claims. This means the final answer may contain claims from different sampled answers. More importantly, we observe that models with better atomic-level calibration (e.g., Qwen2 in Table \ref{tab:reunion}, Appendix \ref{app:applications}) exhibit greater improvements after the reunion process, emphasizing the importance of examining and refining atomic calibration.

\section{Conclusion}

Our main contributions are three-fold: (1) We systematically study \textbf{atomic calibration}, which evaluates confidence calibration at the level of individual atomic claims. Our experiments reveal that models that appear \textit{well-calibrated} at the response level \textit{perform poorly at the atomic level}. (2) To analyze confidence elicitation methods, we categorize them into discriminative and generative methods. We also propose two novel fusion strategies to combine the confidence scores based on confidence agreement. (3) Our atomic-level analysis provides further insights into confidence methods alignment and confidence changes during generation. We find with discriminative methods, models show \textit{decreasing confidence} in atomic facts as generation progresses. In contrast, generative methods show the \textit{lowest average confidence in the middle} of the generation. Last but not least, we demonstrate the utilities of atomic calibration and propose for future research on more fine-grained confidence in long-form generation.

\section*{Limitation}

First, our work primarily focuses on the factuality aspect of LLMs. As mentioned in Section \ref{sec:atomic}, the task $t$ can be various aspects of the quality of a long-form response, such as coherence, creativity, writing style, and more. Unlike previous studies that use the overall quality of long-form responses to evaluate calibration \citep{huang2024calibrating}, we concentrate specifically on factuality in this paper. We argue that the hallucination problem is among the most significant challenges faced by LLMs \citep{zhang2023siren, huang2023survey}.

Second, we test the calibration only on open-source LLMs for two main reasons: (1) After assessing the atomic and macro calibration levels of LLMs, our next step is to adjust the model to better reflect its confidence (\ie for better calibration). Closed-source models are not directly applicable to this calibration process. (2) Our discrimination methods typically require logit access, which is generally unavailable in closed-source models. If logits are accessible, our methods can be directly applied to closed-source models without affecting the atomic calibration process.

Third, in this work, we mainly focus on exploring different confidence elicitation methods and therefore do \textbf{not} apply post-hoc calibration techniques such as histogram binning or temperature scaling. Applying these methods makes it difficult to disentangle improvements due purely to \textit{elicitation} from those due to \textit{recalibration}. To isolate the contribution of our elicitation designs and to avoid conflating them with downstream post-processing effects, we report raw atomic and macro confidence scores, leaving a systematic study of post-hoc techniques to future work.

\section*{Ethics Statement}

Our research adheres to strict ethical standards. We ensured compliance with the licenses of all datasets and models used. No human participants were involved in our experiments. After thorough assessment, we do not anticipate any additional ethical concerns or risks related to our work.

\bibliography{anthology,custom}

\newpage

\appendix
\section*{Appendix}
\label{sec:appendix}

\section{Atomic Calibration Metrics} \label{app:atomic_metrics}

\paragraph{ECE} 
In computing the Expected Calibration Error (ECE), the predictions are sorted and divided into a fixed number of bins $K$. The predicted value of each test instance falls into one of the bins. $ECE$ uses empirical estimates as follows:
$$
ECE=\sum_{i=1}^K P(i) \cdot \left| o_i - e_i \right|,
$$
where $o_i$ is the true fraction of positive instances in bin $i$, $e_i$ is the mean of the post-calibrated probabilities for the instances in bin $i$, and $P(i)$ is the empirical probability (fraction) of all instances that fall into bin $i$. The lower the $ECE$ value, the better a model is calibrated.

When labels are continuous values between 0 and 1, the ECE formulation can be generalized. Instead of binning instances based on binary outcomes, the continuous predictions are grouped into bins according to their predicted probability values. Specifically, the observed calibration error $o_i$ in each bin is the average of the continuous label values for the instances in that bin, and $e_i$ is the mean predicted probability for those instances. This ensures that the calibration error accounts for all possible real-valued outcomes within the range [0,1], providing a more nuanced measure of calibration when the labels are continuous.

\paragraph{Brier Score}  
The Brier score measures the accuracy of probabilistic predictions. In binary classification, it compares the predicted probability of the positive class with the actual binary outcome (0 or 1). The Brier score is defined as:
$$
P = \frac{1}{n} \sum_{i=1}^n \left( \hat{y}_i - y_i \right)^2,
$$
where $\hat{y}_i$ is the predicted probability for instance $i$ and $y_i \in \{0, 1\}$ is the actual binary outcome.

For continuous labels in the range $[0, 1]$, the Brier score can still be used, where $y_i$ is now a continuous value between 0 and 1. In this case, the Brier score becomes equivalent to the mean squared error (MSE) between predicted probabilities and the true values, and minimizing the Brier score for continuous labels is analogous to minimizing MSE. Both metrics aim to reduce the squared differences between predicted and true values, with lower scores indicating better calibration and accuracy.

\paragraph{AUROC} Following \citep{kuhn2022semantic}, AUROC metric is equivalent to the probability that a randomly chosen correct answer has a higher confidence score than a randomly chosen incorrect answer. Higher scores are better for AUROC, and perfect confidence score is 1, while a random confidence measure would be 0.5.

\paragraph{Spearman Correlation} Following \citet{zhang2024luq}, we calculate Spearman Correlation to assess whether samples with higher
factuality have correspondingly higher confidence scores. Compared to Pearson Correlation, it focuses on assessing the rank correlation, is robust to outliers and does not require that data is in normal
distribution.

\section{Statistics in Atomic and Macro Calibration}

To assess the confidence of a model, we generate responses using various questions (e.g., $N$ questions). For each response, a single confidence score is too coarse-grained. Instead, we evaluate the confidence of each atomic claim, with an average of $M$ atomic claims per response. These individual confidences are then aggregated into a response-level confidence score. 

Atomic calibration is computed over $MN$ data points, where $M$ is the average number of atomic claims per response, and $N$ is the number of responses. In contrast, response-level calibration is based on $N$ data points. This distinction highlights the trustworthiness of the model's confidence at both the atomic and response levels, providing a more granular view of its performance.

From the above discussion, it follows that to ensure sufficient data points for atomic calibration, $MN$ must be large. In our datasets, $N$ typically exceeds 1,000, ensuring that $MN$ remains robust even when some responses have only a few calims. 

\begin{figure*}[h!]
    \centering
    \includegraphics[width=0.8\textwidth]{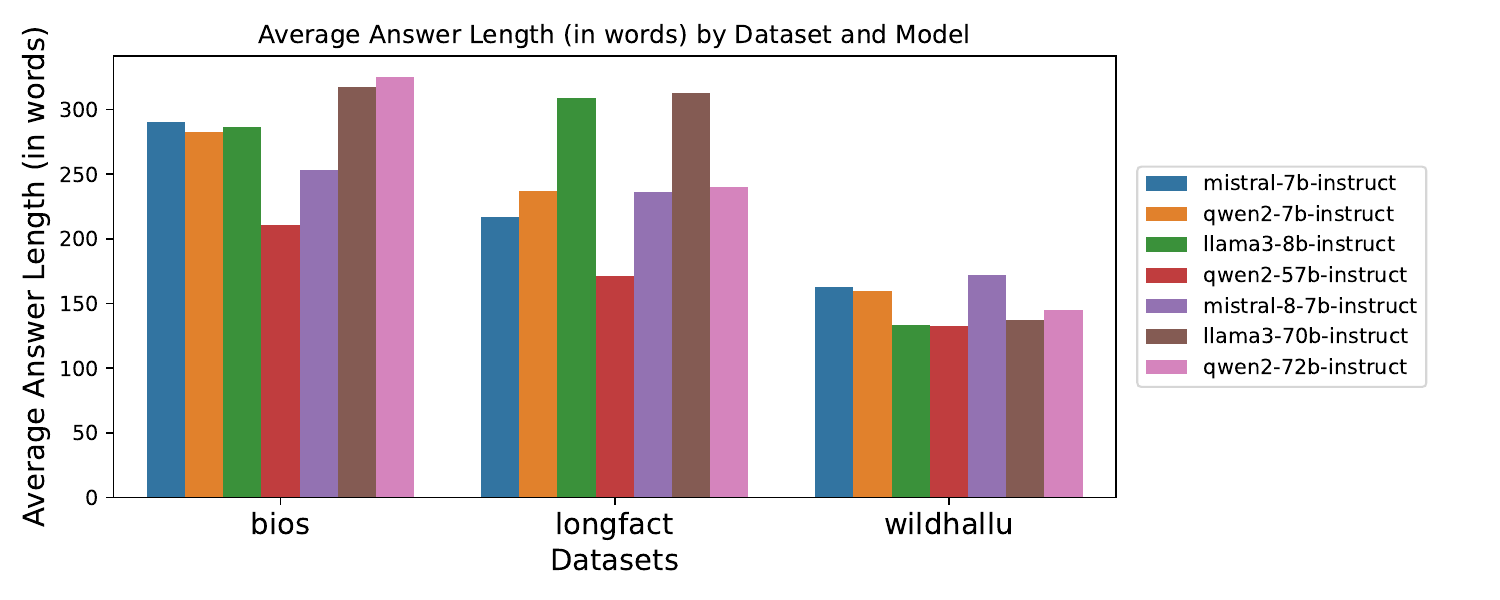}
    \caption{Average answer length (in words) for different models on \texttt{Bios}, \texttt{longfact}, and \texttt{wildhallu}.}
    \label{fig:avg_answer_length}
\end{figure*}

\begin{figure*}[h!]
    \centering
    \includegraphics[width=0.8\textwidth]{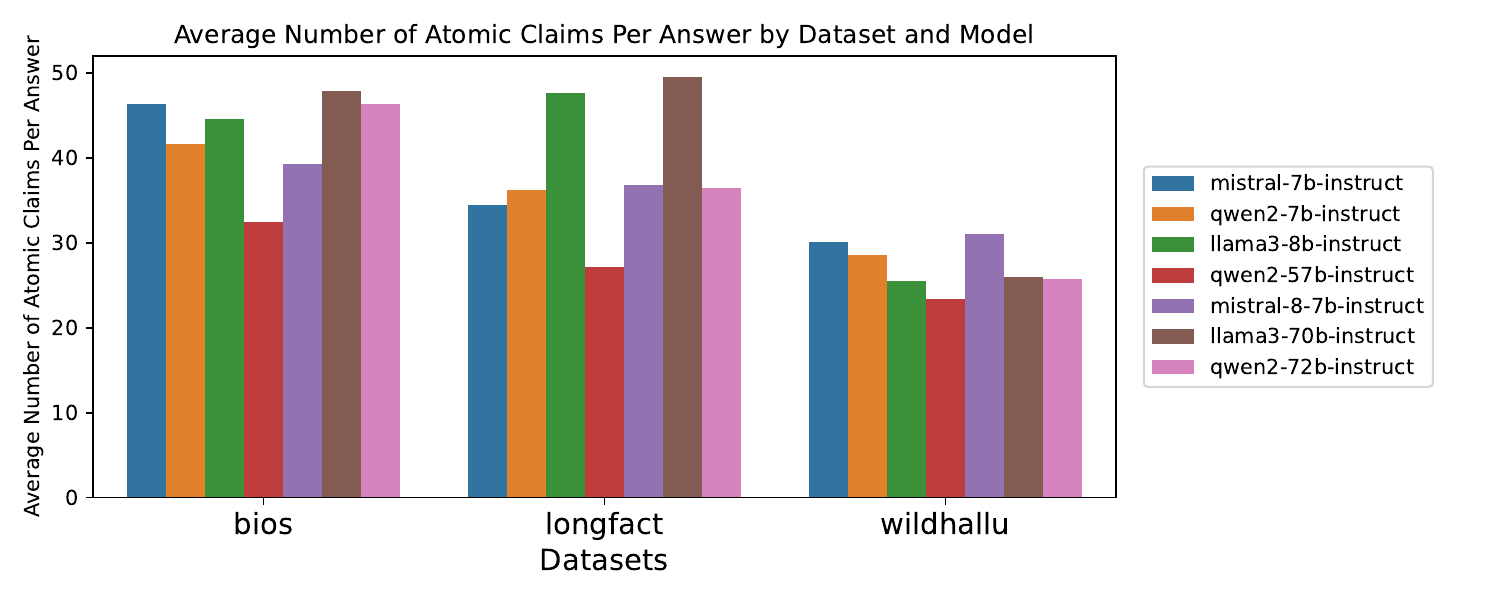}
    \caption{Average number of atomic claims per answer for different models on \texttt{Bios}, \texttt{longfact}, and \texttt{wildhallu}.}
    \label{fig:avg_atomic_fact_length}
\end{figure*}

\begin{figure*}[h!]
    \centering
    \includegraphics[width=0.8\textwidth]{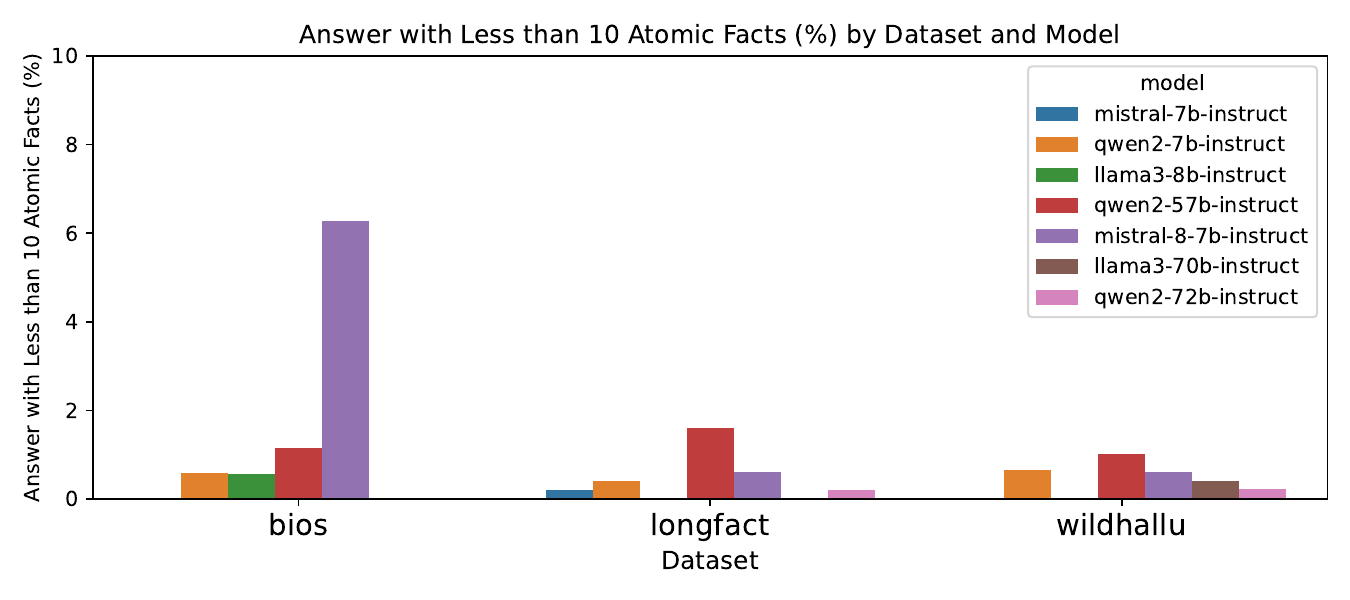}
    \caption{Answer with less than 10 atomic facts (\%) by dataset and model. Notably, \texttt{Mistral-8$\times$7B-Instruct} has 6\% short answers. Human evaluation reveals that these responses are primarily instances where the model refuses to answer.}
    \label{fig:short_atomic_fact}
\end{figure*}

The detailed generation statistics are further illustrated in Figures \ref{fig:avg_answer_length}, \ref{fig:avg_atomic_fact_length}, and \ref{fig:short_atomic_fact}. Figure \ref{fig:avg_answer_length} presents the average answer length, while Figure \ref{fig:avg_atomic_fact_length} shows the average number of atomic claims per answer. Finally, Figure \ref{fig:short_atomic_fact} highlights the percentage of answers containing fewer than 10 atomic facts.

\clearpage
\newpage
\clearpage
\newpage
\section{Applications} \label{app:applications}

\paragraph{Selective Question Answering:} Selective question answering involves setting a confidence threshold, which can be derived from a validation set, to selectively reject questions with low confidence. This approach aims to improve the overall factuality of the responses by eliminating potentially unreliable answers.

Using the \textit{Bios} dataset, we evaluate the performance of three models: Llama3-8B-Instruct, Mistral-7B-Instruct, and Qwen2-7B-Instruct with \textsc{Dis-Gen} and Semantic Entropy (SE). We observe an improvement in overall factuality as we gradually rejected more questions (from 0\% to 10\%). The table below illustrates this trend, highlighting the utility of \textsc{Dis-Gen} in identifying accurate responses and improving selective question answering. Our comparison between \textsc{Dis-Gen} and Semantic Entropy (SE) indicates that \textsc{Dis-Gen} brings more significant improvements in factuality, suggesting that better calibration methods can substantially enhance the results of selective question answering.

\paragraph{LLM Ensemble:} In the \textit{LLM Ensemble} method, we use three models to generate answers to the same question and select the response from the model with the highest confidence. This approach aims to enhance factuality by leveraging the strengths of each model. The \textbf{Answer Distribution (AD)} shows the proportion of the final response contributed by each model, highlighting the benefit of ensemble methods. The table below presents the results of applying this method on the Bios and WildHallu datasets, comparing two different selection strategies: \textsc{DIS-Gen} and \textsc{SE}.

The results demonstrate that the ensemble method with \textsc{DIS-Gen} significantly improves factuality. For instance, in the Bios dataset, the factuality score increases from 0.475 to 0.556, and in the WildHallu dataset, it increases from 0.655 to 0.752. In contrast, using \textsc{SE} results in no improvements, with factuality scores even lower than the best individual model (0.484 vs 0.502 and 0.671 vs 0.701) for the Bios and WildHallu datasets, respectively. These findings suggest that \textbf{ensembling does not always guarantee better results}, and the selection strategy, such as DIS-Gen, plays a crucial role in improving factuality.

\paragraph{Atomic Reunion:} 
In \textit{Atomic Reunion}, for each question, we begin by sampling the model's output five times (this is also what we need to calculate \textsc{Gen-Dis}. These outputs are then broken down into atomic claims, which are individual, verifiable statements. Each claim is evaluated for confidence, and only those with a high confidence level are retained. Subsequently, we prompt a LLM, such as GPT-4o, to reassemble the selected atomic claims into a cohesive and factually accurate response.

This method seeks to enhance factuality by utilizing smaller, more manageable pieces of information, allowing the LLM to generate a more reliable answer by combining only high-confidence atomic claims. The table below presents the factuality scores of the new answers generated through the Atomic Reunion approach. As observed, this approach leads to a significant improvement in factuality compared to the baseline models. 

\begin{table*}[h]
\centering
\footnotesize
\begin{tabular}{cccc}
\toprule
\textbf{Refuse Rate} & \textbf{Llama3-8B-Instruct} & \textbf{Mistral-7B-Instruct} & \textbf{Qwen2-7B-Instruct} \\
\midrule
\rowcolor[gray]{0.9}
\multicolumn{4}{l}{\textbf{SE}} \\
\midrule
0\%  & 0.475 & 0.403 & 0.502 \\
5\%  & 0.479 & 0.407 & 0.506 \\
7.5\% & 0.483 & 0.411 & 0.511 \\
10\% & 0.485 & 0.416 & 0.518 \\
\midrule
\rowcolor[gray]{0.9}
\multicolumn{4}{l}{\textbf{DIS-Gen}} \\
\midrule

0\%  & 0.475 & 0.403 & 0.502 \\
5\%  & 0.496 & 0.419 & 0.517 \\
7.5\% & 0.511 & 0.438 & 0.533 \\
10\% & \textbf{0.528} & \textbf{0.465} & \textbf{0.557} \\
\bottomrule
\end{tabular}
\caption{Factuality Scores with Varying Refuse Rates for Selective Question Answering}
\end{table*}

\begin{table*}[h!]
\centering
\footnotesize
\begin{tabular}{lccccc}
\toprule
\textbf{Model} & \multicolumn{2}{c}{\textbf{Bios}} & \multicolumn{2}{c}{\textbf{WildHallu}} \\
\midrule
 & \textbf{Factuality Scores} & \textbf{Answer Distribution} & \textbf{Factuality Scores} & \textbf{Answer Distribution} \\
\midrule
\rowcolor[gray]{0.9}
\multicolumn{5}{l}{\textbf{DIS-Gen}} \\
\midrule
Llama3-8B-Instruct  & 0.475 & 31\% & 0.655 & 29\% \\
Mistral-7B-Instruct  & 0.403 & 23\% & 0.631 & 27\% \\
Qwen2-7B-Instruct     & 0.502 & 46\% & 0.701 & 44\% \\
\textbf{Ensemble}    & \textbf{0.556} & / & \textbf{0.752} & / \\
\midrule
\rowcolor[gray]{0.9}
\multicolumn{5}{l}{\textbf{SE}} \\
\midrule
Llama3-8B-Instruct  & 0.475 & 18\% & 0.655 & 15\% \\
Mistral-7B-Instruct  & 0.403 & 37\% & 0.631 & 44\% \\
Qwen2-7B-Instruct     & \textbf{0.502} & 45\% & \textbf{0.701} & 41\% \\
\textbf{Ensemble}    & 0.484 & / & 0.671 & / \\
\bottomrule
\end{tabular}
\caption{LLM Ensemble Results on Bios and WildHallu Datasets}
\end{table*}

\begin{table*}[h!]
\centering
\footnotesize
\begin{tabular}{cccccc}
\toprule
\textbf{Model} & \textbf{Before Atomic Reunion} & \textbf{After Atomic Reunion} \\
\midrule
\rowcolor[gray]{0.9}
\multicolumn{3}{l}{\textbf{DIS-Gen}} \\
\midrule
Llama3-8B-Instruct  & 0.475 & 0.501 \\
Mistral-7B-Instruct  & 0.403 & 0.441 \\
Qwen2-7B-Instruct     & 0.502 & \textbf{0.575} \\
\bottomrule
\end{tabular}
\caption{Factuality Scores Before and After Atomic Reunion using DIS-Gen}
\label{tab:reunion}
\end{table*}

\section{Reliability of Atomic Facts Generation and Verification}

The processes of atomic fact generation and verification have been extensively studied and validated in prior work \cite{min-etal-2023-factscore,wei2024longfact,zhao2024wildhallu}. For instance, FActScore \cite{min-etal-2023-factscore} reports an error rate of 2\%. In this work, we leverage their pipeline while employing stronger models from GPT-3.5 and GPT-4o, to further enhance performance.

We, the authors, conducted additional tests comparing GPT’s atomic decompositions with ground-truth manual segmentations. We manually selected 30 samples for this evaluation. The results of our assessment are as follows:

\begin{itemize}
    \item \textbf{Consistency:} Over 10 trials, GPT demonstrated a high inter-run consistency of 95\%, indicating stable and repeatable outcomes.
    \item \textbf{Error Rate:} The error rate, which includes missing or overly segmented claims and misclassification of factuality, was measured at 6.4\%. This error rate is manageable within the context of our calibration framework, suggesting that the model’s atomic fact generation is reliable.
\end{itemize}

\section{Experiment Details} \label{app:exp}

We use vLLM \citep{kwon2023efficient} for our LLM inference tasks, with the following parameters: temperature = 1, top-$p$ = 0.95, and a maximum output of 512 tokens. For discriminative confidence elicitation methods, we set the temperature to 0 and only consider the top 10 logits. For generative methods, we use $N=20$ samples. The experiments are conducted on A100-SXM-40GB GPUs. Running the discriminative methods takes 30 minutes for 500 samples, while the generative methods take 1.3 hours for the same number of samples. We use GPT-4o as the auxiliary model for generating atomic claims and fact-checking the LLM.

\onecolumn
\section{Confidence Fusion Results} \label{app:fusions}

\begin{table*}[h!]
\centering
\footnotesize \centering
\begin{tabular}{lccccccccc}
\toprule
 & \multicolumn{3}{c}{\textbf{Bios}} & \multicolumn{3}{c}{\textbf{LongFact}} & \multicolumn{3}{c}{\textbf{WildHallu}} \\
\cmidrule(lr){2-4}  \cmidrule(lr){5-7} \cmidrule(lr){8-10}
\multicolumn{1}{c}{} & \multicolumn{1}{c}{ECE $\downarrow$} & \multicolumn{1}{c}{BS $\downarrow$} & \multicolumn{1}{c}{AUROC $\uparrow$} & \multicolumn{1}{c}{ECE $\downarrow$} & \multicolumn{1}{c}{BS $\downarrow$} & \multicolumn{1}{c}{AUROC $\uparrow$} & \multicolumn{1}{c}{ECE $\downarrow$} & \multicolumn{1}{c}{BS $\downarrow$} & \multicolumn{1}{c}{AUROC $\uparrow$} \\
\midrule
\textsc{Gen-Binary} & 13.7 & 19.0 & 81.9 & 8.4 & 11.5 & 80.1 & 12.7 & 17.0 & 81.3 \\
\textsc{Dis-Rating} & 44.5 & 42.5 & 65.0 & 10.0 & 14.2 & 67.9 & 19.7 & 23.9 & 68.1 \\
\textsc{Dis-Context} & 24.8 & 26.0 & 77.5 & 15.7 & 16.1 & 75.3 & 20.6 & 21.7 & 79.8 \\

\addlinespace[0.5ex]
\cdashline{1-10}
\addlinespace[0.5ex]

\texttt{MinConf} & 14.1 & 18.3 & 82.0 & 8.6 & 12.7 & 80.7 & 7.6 & 16.0 & 83.2 \\
\texttt{HMean} & 14.3 & 18.3 & 82.1 & 7.6 & 12.2 & 81.0 & 11.5 & 16.6 & 83.4 \\
\texttt{ProdConf} & 14.2 & 18.3 & 82.3 & 9.5 & 13.0 & 81.0 & 7.9 & 15.9 & 83.5 \\
\texttt{WAvg} & 10.6 & 16.6 & 84.7 & \textbf{5.5} & 10.7 & \textbf{82.1} & 12.3 & 16.4 & \textbf{84.4} \\

\addlinespace[0.5ex]
\cdashline{1-10}
\addlinespace[0.5ex]

\texttt{AdjustedAlpha} & \textbf{9.8} & 16.7 & \textbf{85.0} & 5.8 & \textbf{10.5} & 81.8 & \textbf{6.5} & 15.2 & 84.0 \\
\texttt{DampedFusion} & 10.2 & \textbf{16.5} & 84.6 & 5.9 & 10.9 & 81.9 & 7.1 & 15.4 & 83.8 \\

\bottomrule
\end{tabular}
\caption{Atomic calibration results of confidence fusion strategies for Mistral-7B-Instruct. The fusion results are based on \textsc{Gen-Binary} and \textsc{Dis-Context}.}
\end{table*}

\begin{table*}[h!]
\centering
\footnotesize \centering
\begin{tabular}{lccccccccc}
\toprule
 & \multicolumn{3}{c}{\textbf{Bios}} & \multicolumn{3}{c}{\textbf{LongFact}} & \multicolumn{3}{c}{\textbf{WildHallu}} \\
\cmidrule(lr){2-4}  \cmidrule(lr){5-7} \cmidrule(lr){8-10}
\multicolumn{1}{c}{} & \multicolumn{1}{c}{ECE $\downarrow$} & \multicolumn{1}{c}{BS $\downarrow$} & \multicolumn{1}{c}{AUROC $\uparrow$} & \multicolumn{1}{c}{ECE $\downarrow$} & \multicolumn{1}{c}{BS $\downarrow$} & \multicolumn{1}{c}{AUROC $\uparrow$} & \multicolumn{1}{c}{ECE $\downarrow$} & \multicolumn{1}{c}{BS $\downarrow$} & \multicolumn{1}{c}{AUROC $\uparrow$} \\
\midrule
\textsc{Gen-Binary} & 10.9 & 16.7 & 83.8 & 6.3 & 9.9 & 81.9 & 9.5 & 14.0 & 82.5 \\
\textsc{Dis-Rating} & 41.5 & 39.7 & 64.2 & 3.5 & 11.7 & 62.6 & 8.2 & 18.1 & 70.4 \\
\textsc{Dis-Context} & 26.5 & 28.3 & 75.5 & 13.9 & 14.8 & 77.9 & 17.2 & 19.4 & 81.2 \\

\addlinespace[0.5ex]
\cdashline{1-10}
\addlinespace[0.5ex]

\texttt{MinConf} & 11.3 & 16.9 & 82.4 & 6.3 & 10.3 & 80.5 & 5.0 & 13.8 & 82.6 \\
\texttt{HMean} & 11.1 & 16.9 & 82.7 & 2.7 & 9.6 & 81.7 & 4.9 & 13.6 & 83.8 \\
\texttt{ProdConf} & 12.4 & 17.0 & 83.1 & 8.3 & 10.6 & 81.7 & 7.2 & 13.7 & 83.9 \\
\texttt{WAvg} & 10.7 & 15.9 & \textbf{84.8} & \textbf{2.6} & 9.2 & 82.8 & 6.8 & 13.5 & 84.3 \\

\addlinespace[0.5ex]
\cdashline{1-10}
\addlinespace[0.5ex]

\texttt{AdjustedAlpha} & \textbf{8.9} & 16.0 & 84.6 & 2.9 & \textbf{9.1} & 82.6 & \textbf{4.5} & \textbf{13.2} & 84.1 \\
\texttt{DampedFusion} & 10.2 & \textbf{15.8} & 84.5 & \textbf{2.6} & 9.3 & \textbf{82.9} & 5.2 & 13.3 & \textbf{84.4} \\

\bottomrule
\end{tabular}
\caption{Atomic calibration results of confidence fusion strategies for Qwen2-7B-Instruct. The fusion results are based on \textsc{Gen-Binary} and \textsc{Dis-Context}.}
\end{table*}



\clearpage
\section{Confidence Alignment} \label{app:conf_align}

\captionsetup{font=small}
\newcommand{\modelone}{Mistral-7B-Instruct}
\newcommand{\modeltwo}{Qwen2-7B-Instruct}
\newcommand{\modelthree}{Llama3-8B-Instruct}
\newcommand{\datasetone}{Bios}
\newcommand{\datasettwo}{LongFact}
\newcommand{\datasetthree}{WildHallu}

\begin{figure*}[h!]
    \centering
    \begin{subfigure}[b]{0.3\textwidth}
        \centering
        \includegraphics[width=\textwidth]{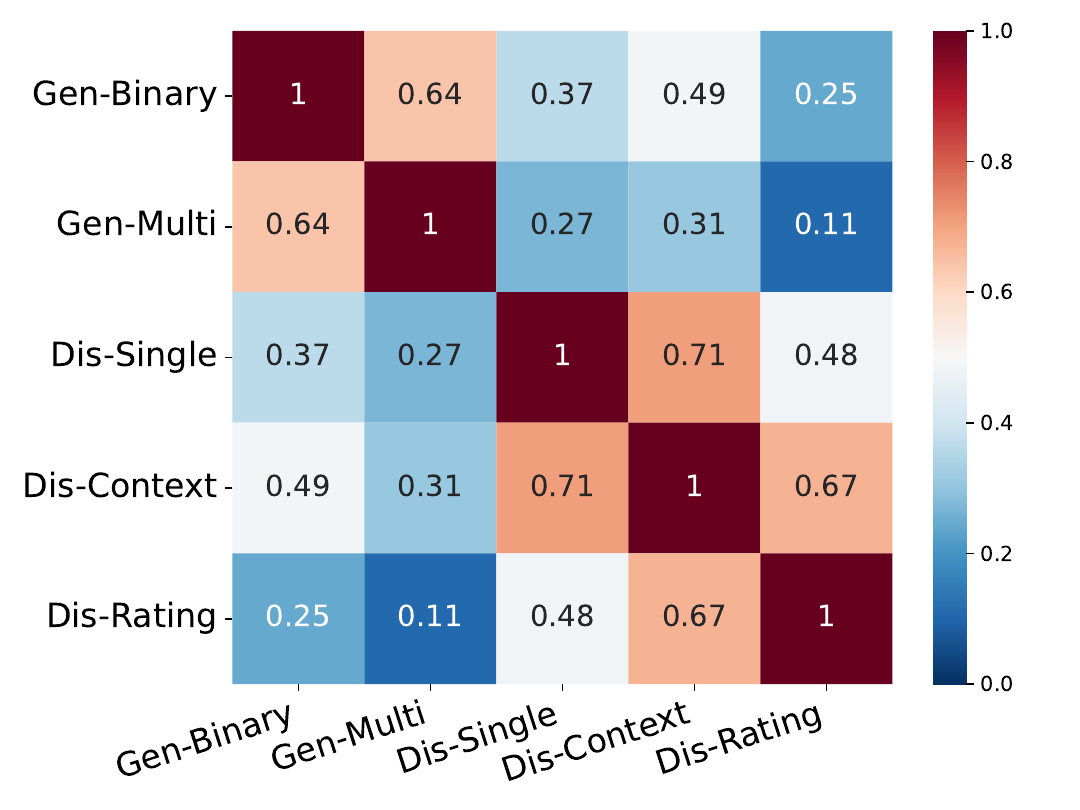}
        \caption{\modelone\ : atomic}
        \label{fig:\modelone_\datasetone_atomic}
    \end{subfigure}
    \begin{subfigure}[b]{0.3\textwidth}
        \centering
        \includegraphics[width=\textwidth]{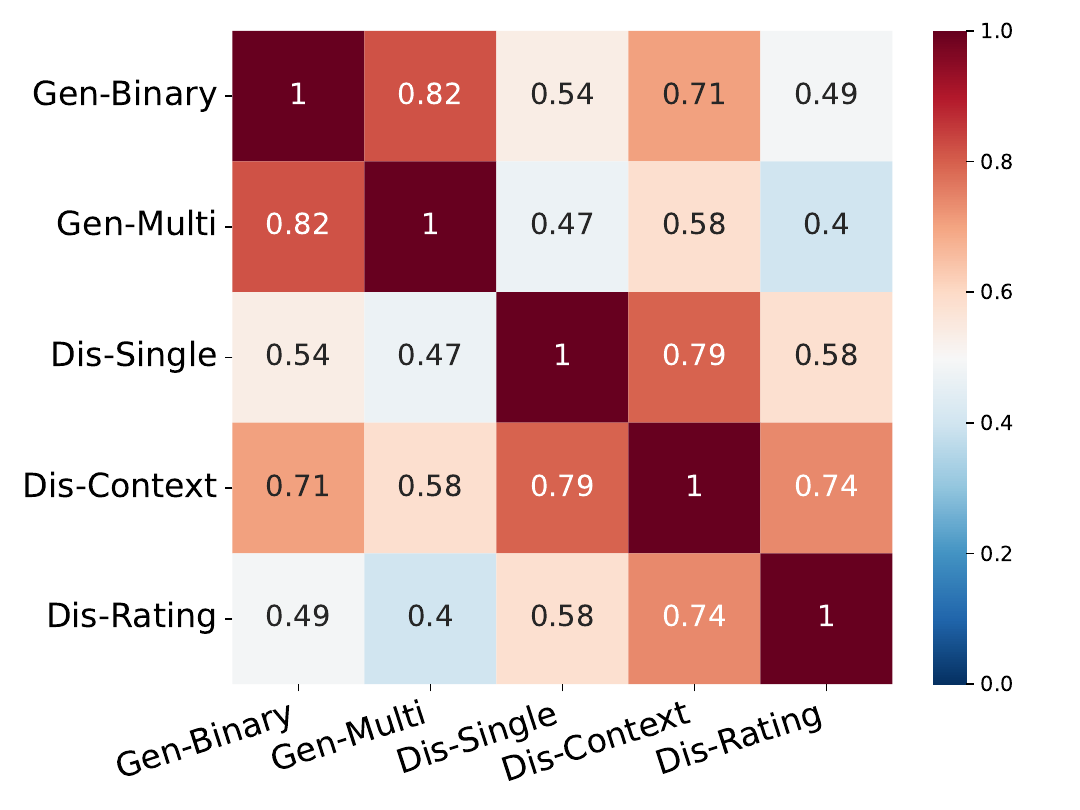}
        \caption{\modelone\ : response}
        \label{fig:\modelone_\datasetone_response}
    \end{subfigure}
    \begin{subfigure}[b]{0.3\textwidth}
        \centering
        \includegraphics[width=\textwidth]{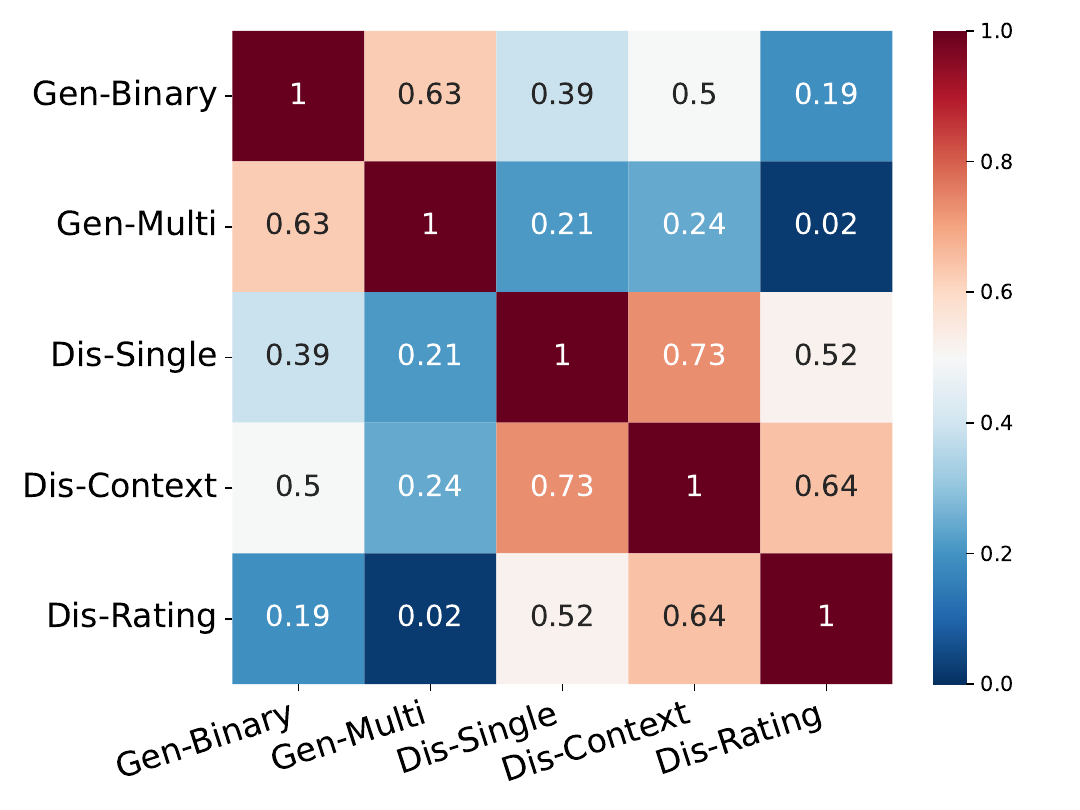}
        \caption{\modeltwo\ : atomic}
        \label{fig:\modeltwo_\datasetone_atomic}
    \end{subfigure}
    \begin{subfigure}[b]{0.3\textwidth}
        \centering
        \includegraphics[width=\textwidth]{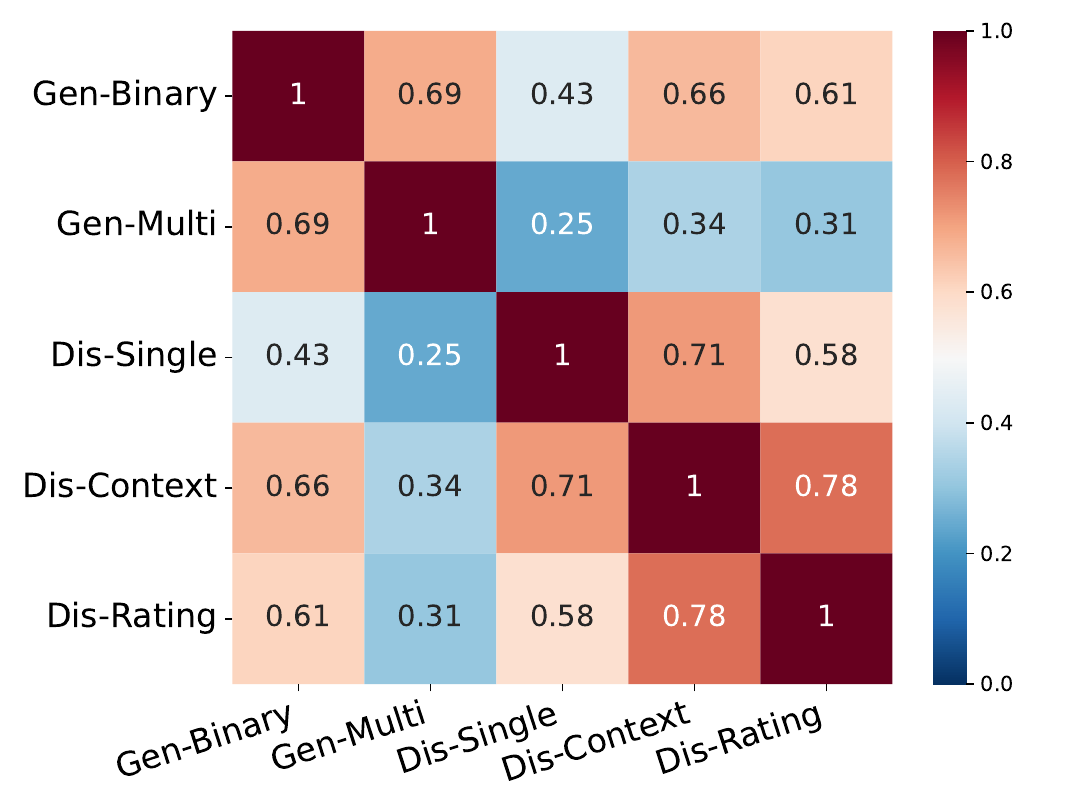}
        \caption{\modeltwo\ : response}
        \label{fig:\modeltwo_\datasetone_response}
    \end{subfigure}
    \begin{subfigure}[b]{0.3\textwidth}
        \centering
        \includegraphics[width=\textwidth]{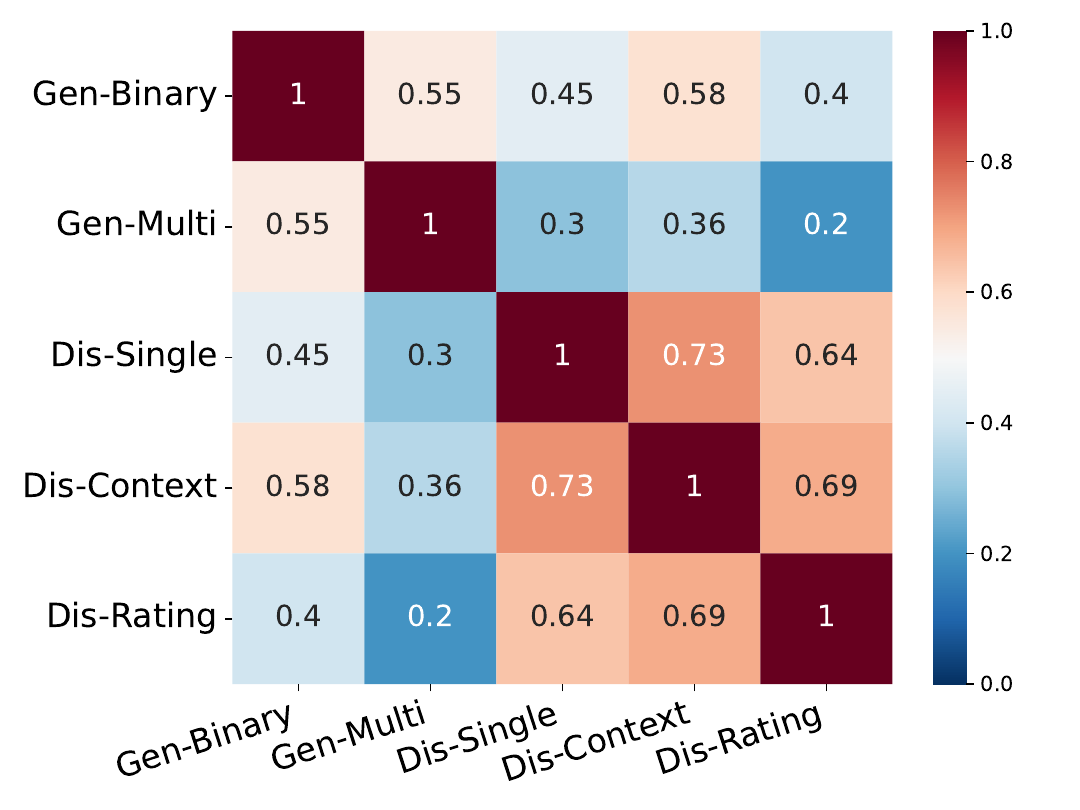}
        \caption{\modelthree\ : atomic}
        \label{fig:\modelthree_\datasetone_atomic}
    \end{subfigure}
    \begin{subfigure}[b]{0.3\textwidth}
        \centering
        \includegraphics[width=\textwidth]{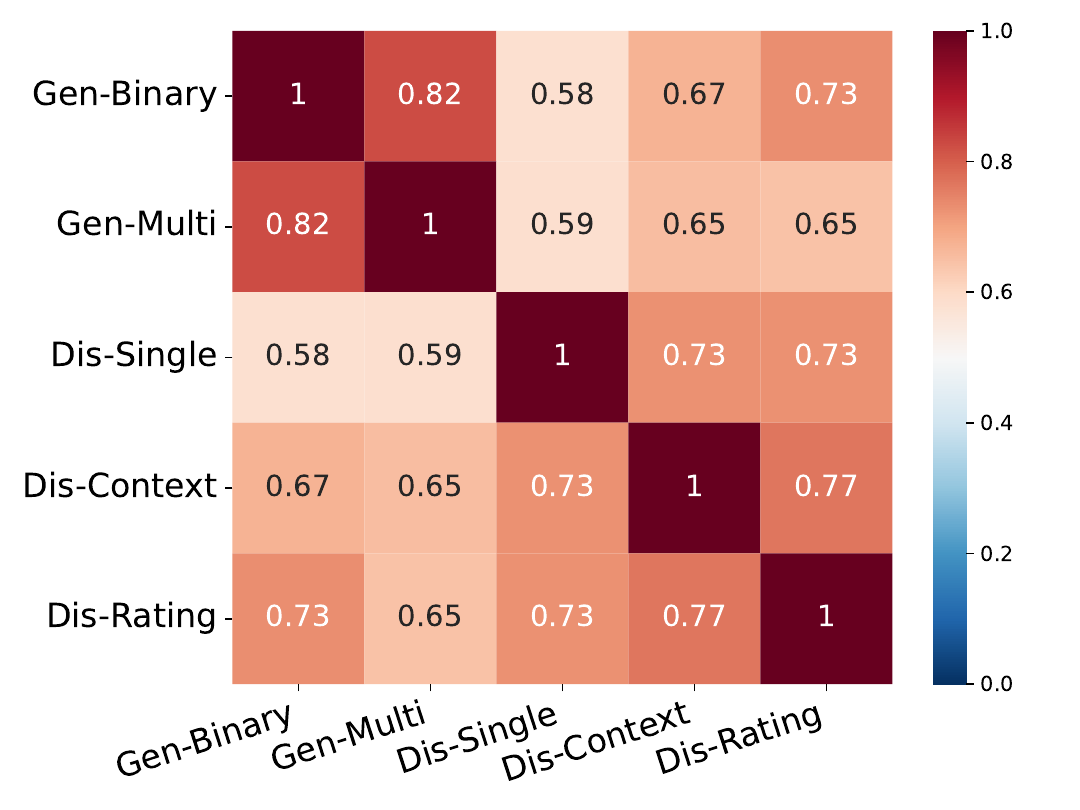}
        \caption{\modelthree\ : response}
        \label{fig:\modelthree_\datasetone_response}
    \end{subfigure}
    \caption{Heatmaps comparing the Spearman correlation between generative and discriminative confidence elicitation methods for \textit{\datasetone}. Results shown for \modelone, \modeltwo, and \modelthree.}
    \label{fig:all_models_\datasetone}
\end{figure*}

\begin{figure*}[h!]
    \centering
    \begin{subfigure}[b]{0.3\textwidth}
        \centering
        \includegraphics[width=\textwidth]{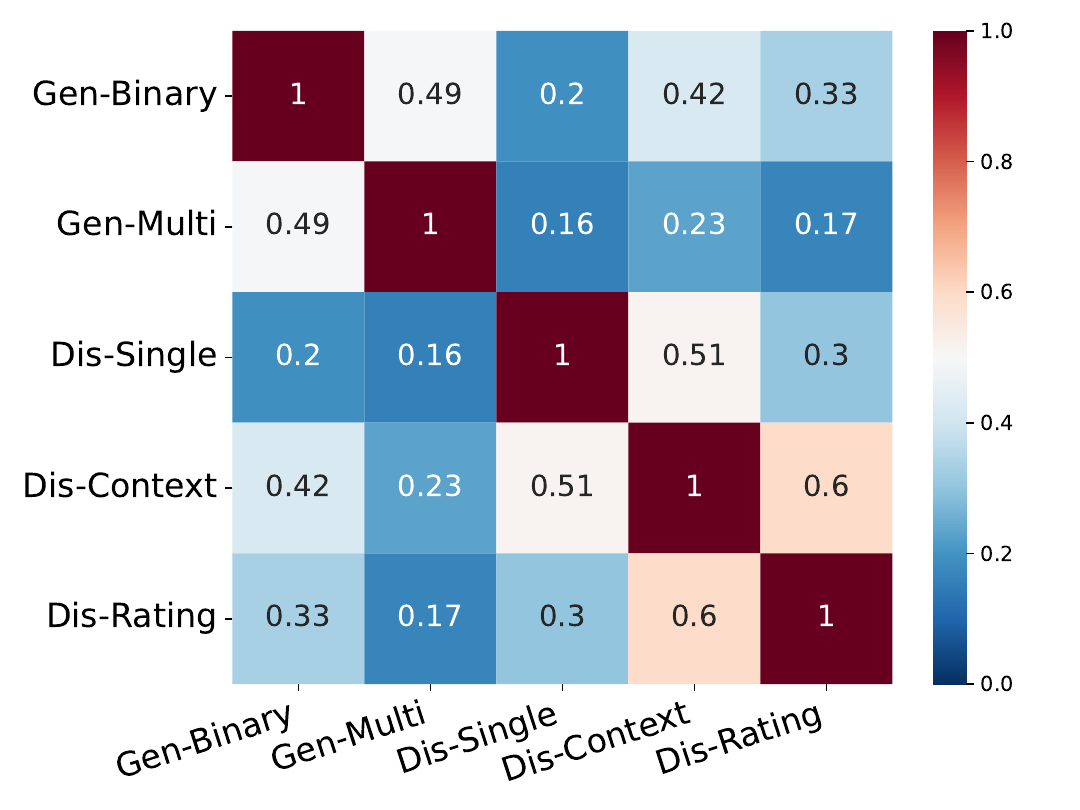}
        \caption{\modelone\ : atomic}
        \label{fig:\modelone_\datasettwo_atomic}
    \end{subfigure}
    \begin{subfigure}[b]{0.3\textwidth}
        \centering
        \includegraphics[width=\textwidth]{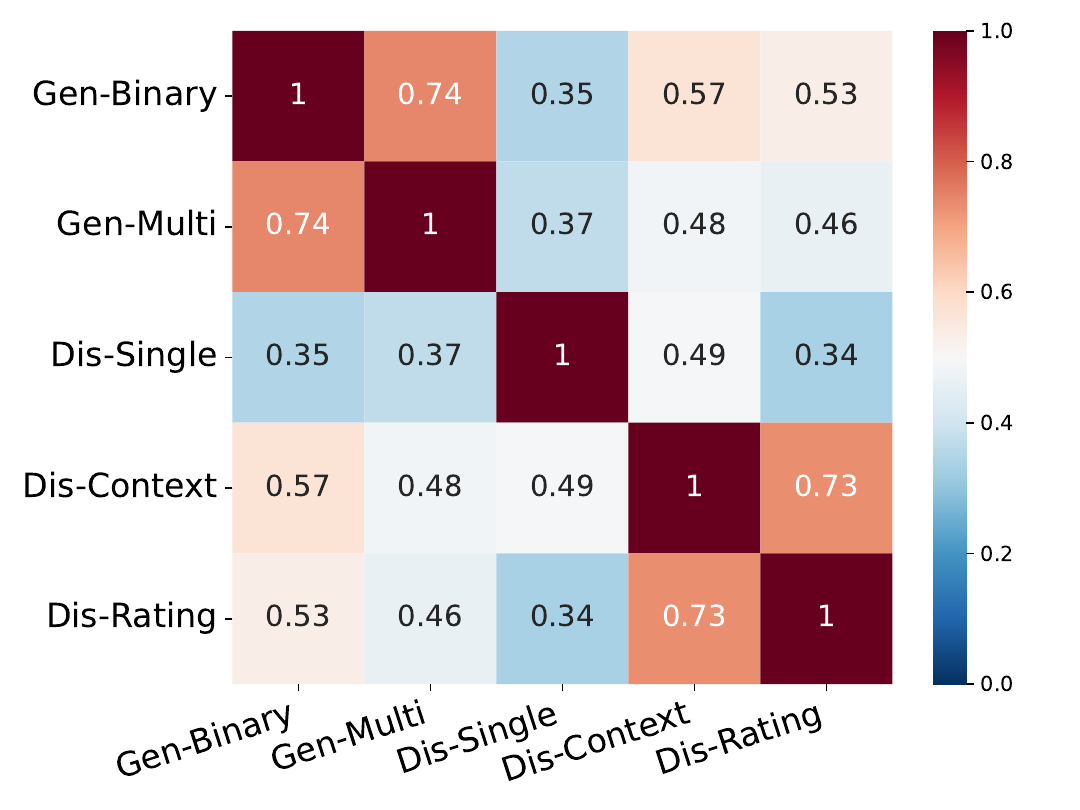}
        \caption{\modelone\ : response}
        \label{fig:\modelone_\datasettwo_response}
    \end{subfigure}
    \begin{subfigure}[b]{0.3\textwidth}
        \centering
        \includegraphics[width=\textwidth]{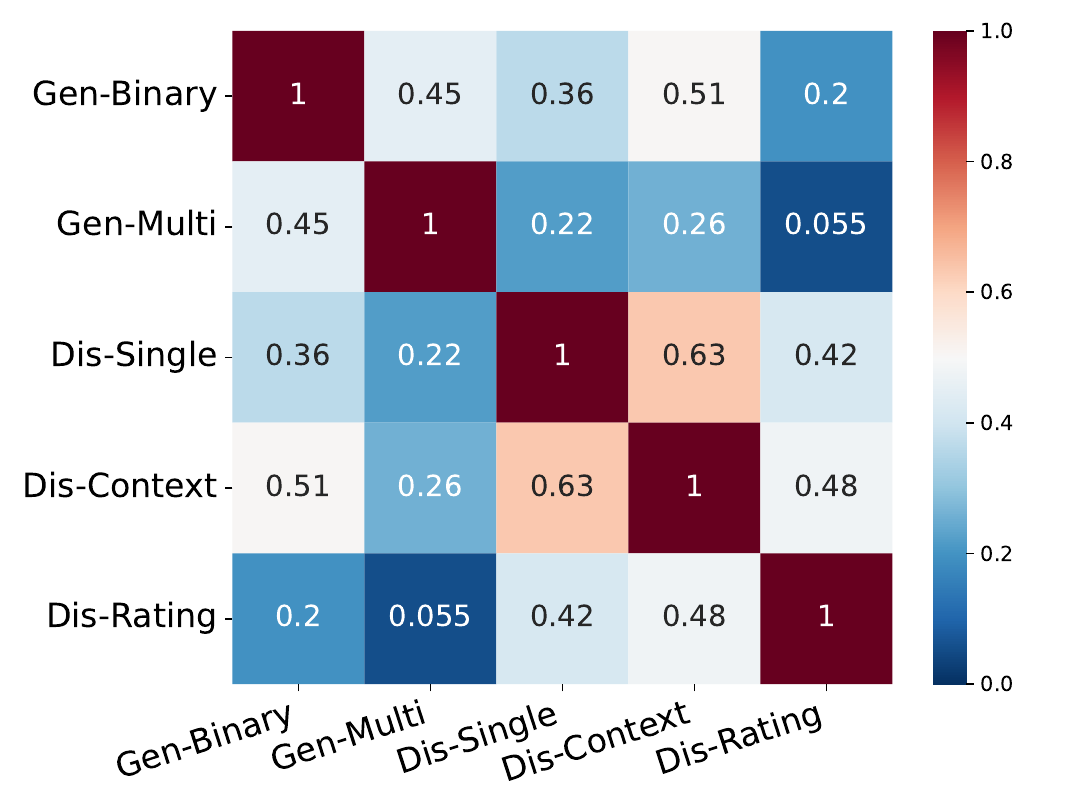}
        \caption{\modeltwo\ : atomic}
        \label{fig:\modeltwo_\datasettwo_atomic}
    \end{subfigure}
    \begin{subfigure}[b]{0.3\textwidth}
        \centering
        \includegraphics[width=\textwidth]{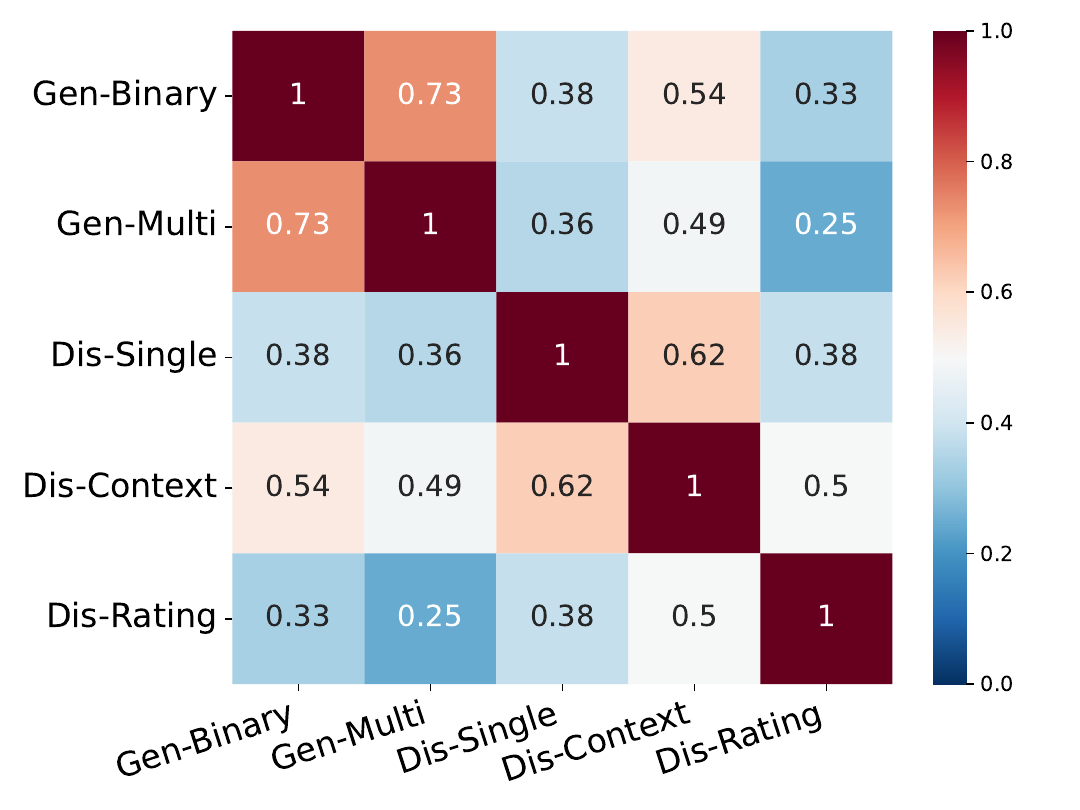}
        \caption{\modeltwo\ : response}
        \label{fig:\modeltwo_\datasettwo_response}
    \end{subfigure}
    \begin{subfigure}[b]{0.3\textwidth}
        \centering
        \includegraphics[width=\textwidth]{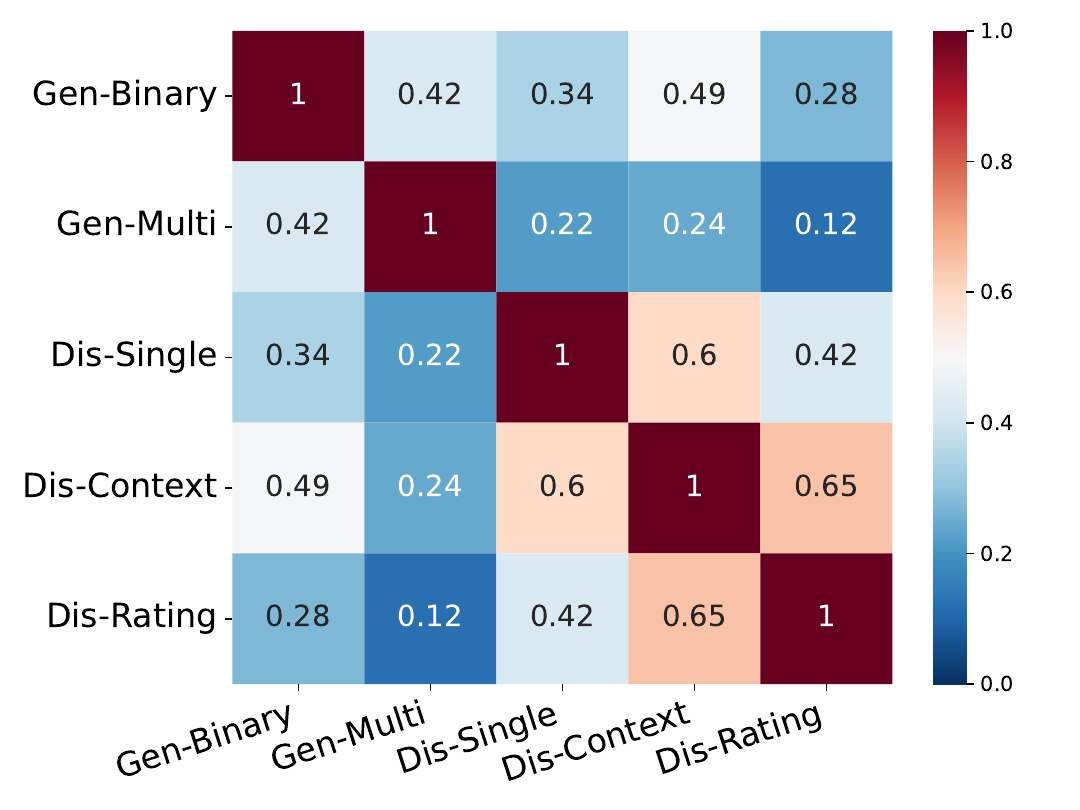}
        \caption{\modelthree\ : atomic}
        \label{fig:\modelthree_\datasettwo_atomic}
    \end{subfigure}
    \begin{subfigure}[b]{0.3\textwidth}
        \centering
        \includegraphics[width=\textwidth]{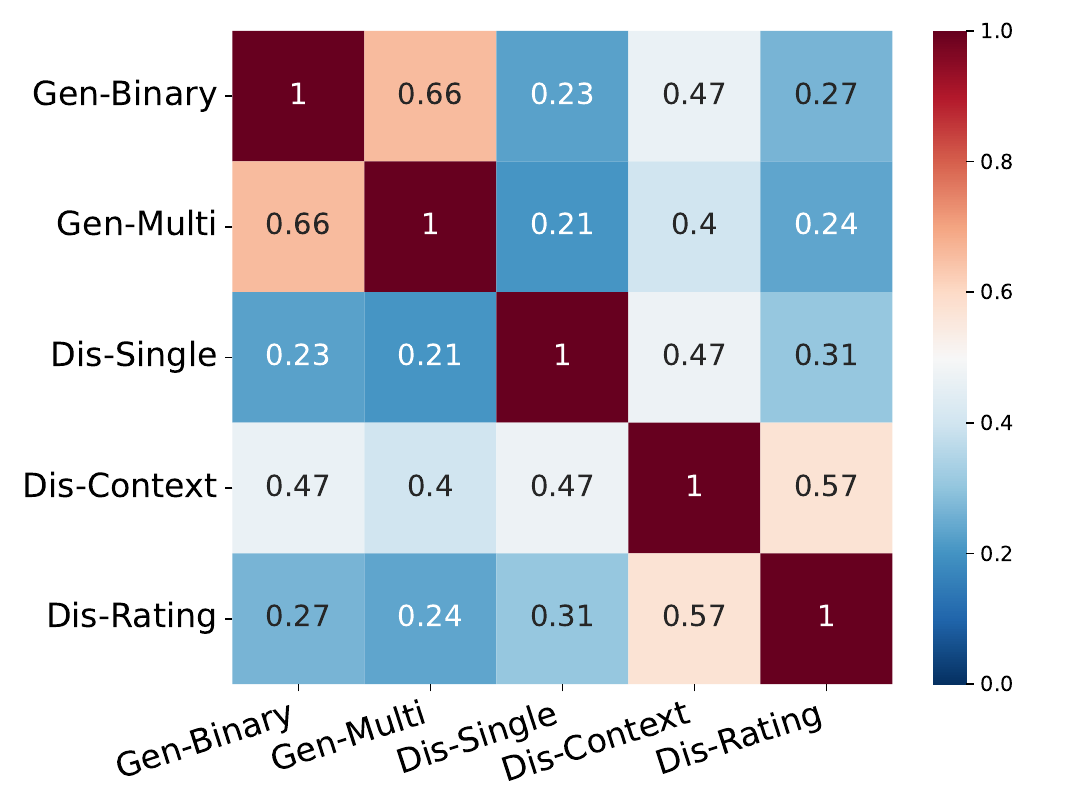}
        \caption{\modelthree\ : response}
        \label{fig:\modelthree_\datasettwo_response}
    \end{subfigure}
    \caption{Heatmaps comparing the Spearman correlation between generative and discriminative confidence elicitation methods for \textit{\datasettwo}. Results shown for \modelone, \modeltwo, and \modelthree.}
    \label{fig:all_models_\datasettwo}
\end{figure*}

\begin{figure*}[h!]
    \centering
    \begin{subfigure}[b]{0.3\textwidth}
        \centering
        \includegraphics[width=\textwidth]{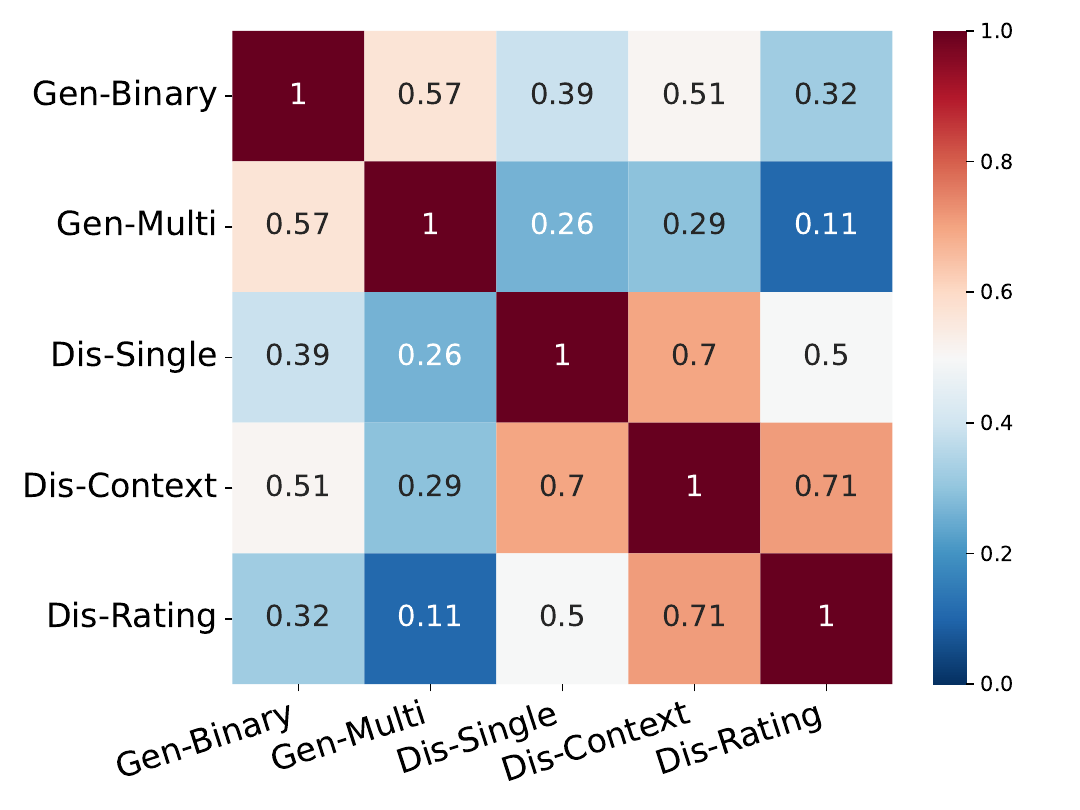}
        \caption{\modelone\ : atomic}
        \label{fig:\modelone_\datasetthree_atomic}
    \end{subfigure}
    \begin{subfigure}[b]{0.3\textwidth}
        \centering
        \includegraphics[width=\textwidth]{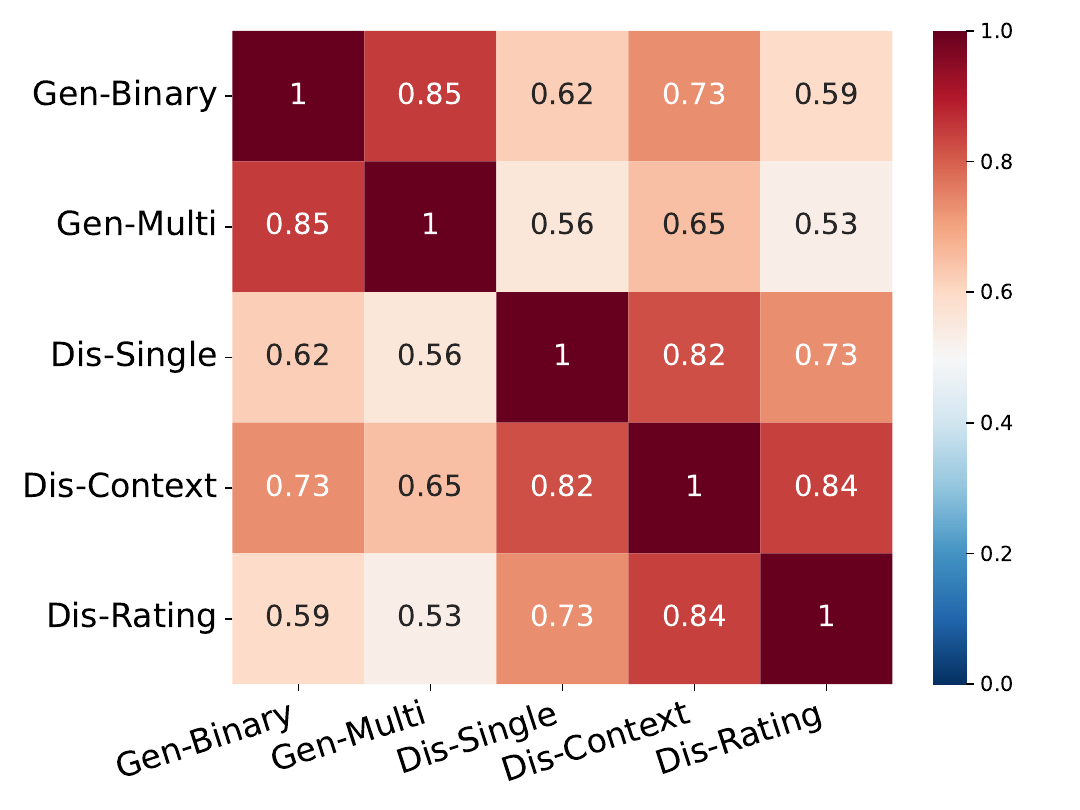}
        \caption{\modelone\ : response}
        \label{fig:\modelone_\datasetthree_response}
    \end{subfigure}
    \begin{subfigure}[b]{0.3\textwidth}
        \centering
        \includegraphics[width=\textwidth]{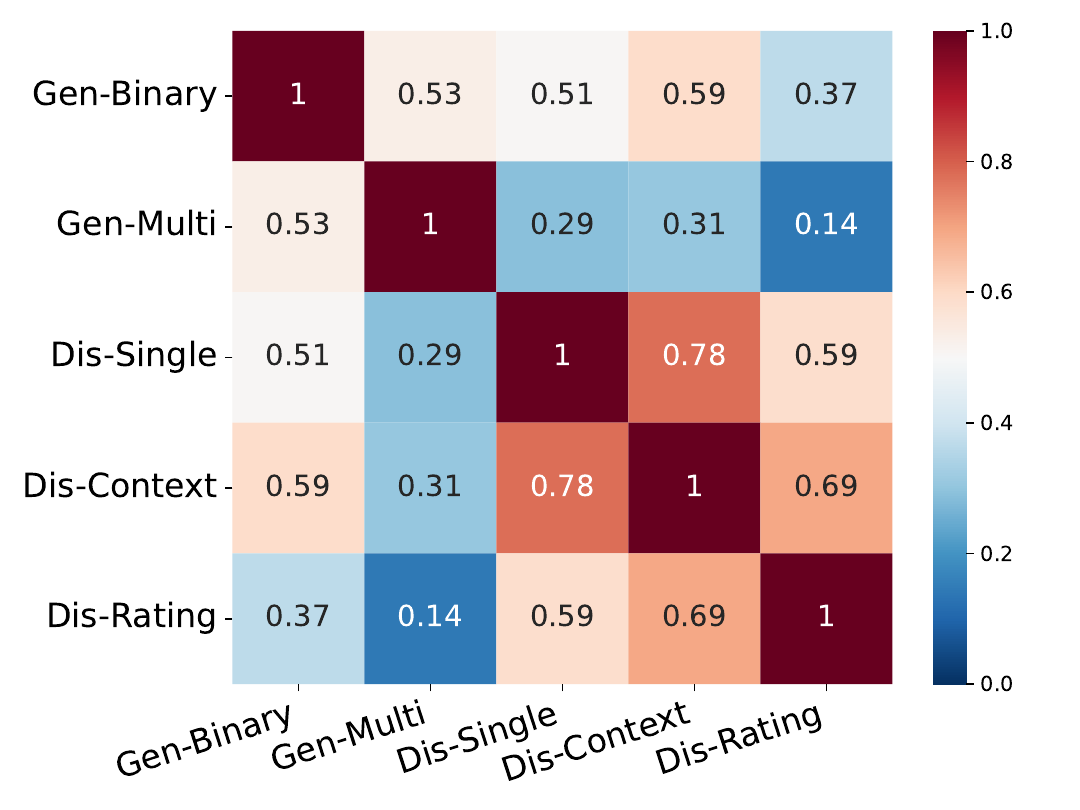}
        \caption{\modeltwo\ : atomic}
        \label{fig:\modeltwo_\datasetthree_atomic}
    \end{subfigure}
    \begin{subfigure}[b]{0.3\textwidth}
        \centering
        \includegraphics[width=\textwidth]{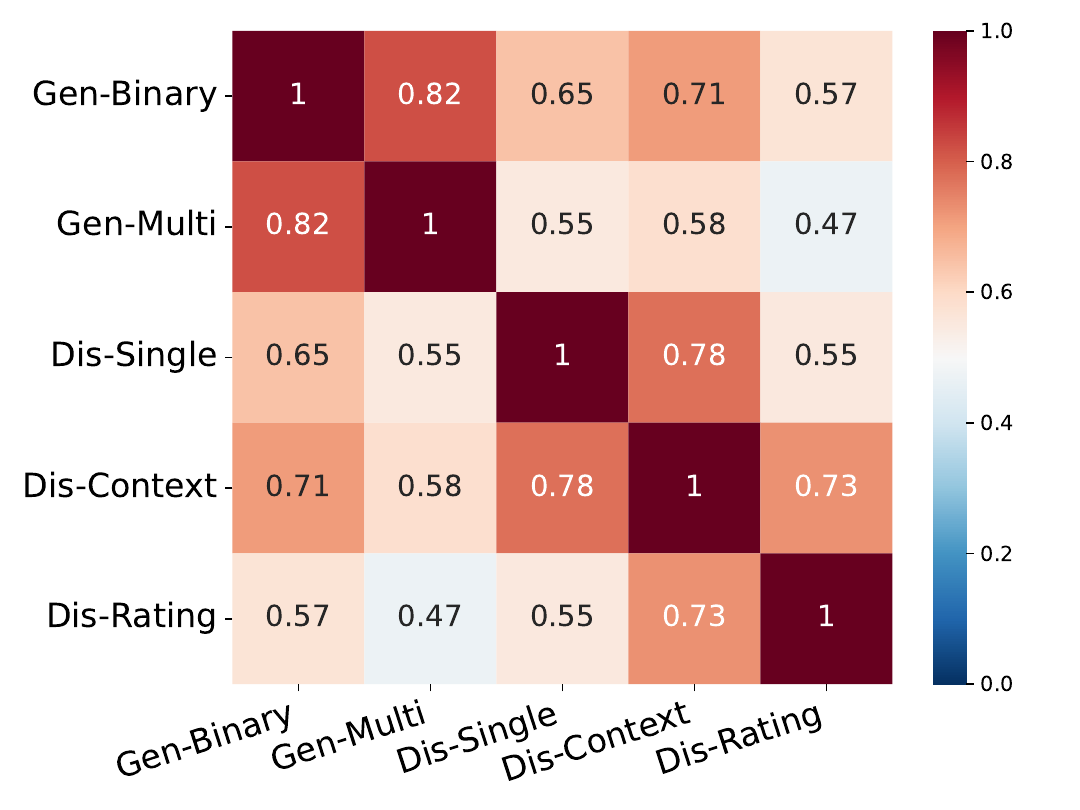}
        \caption{\modeltwo\ : response}
        \label{fig:\modeltwo_\datasetthree_response}
    \end{subfigure}
    \begin{subfigure}[b]{0.3\textwidth}
        \centering
        \includegraphics[width=\textwidth]{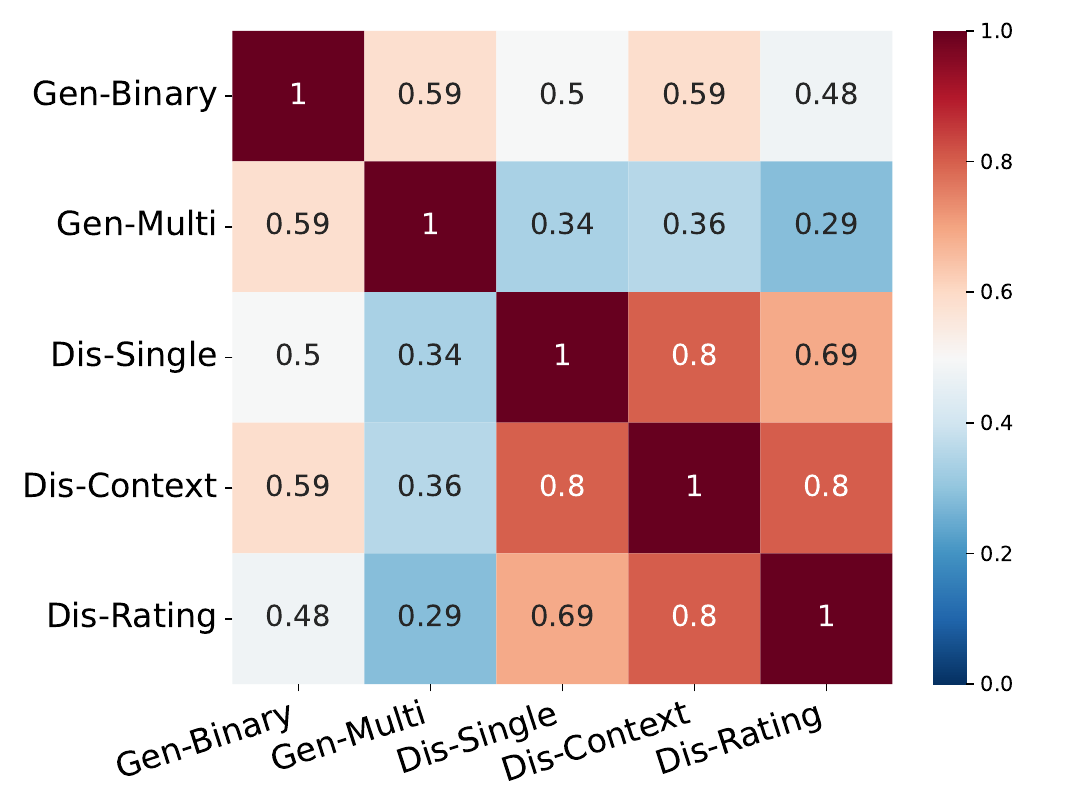}
        \caption{\modelthree\ : atomic}
        \label{fig:\modelthree_\datasetthree_atomic}
    \end{subfigure}
    \begin{subfigure}[b]{0.3\textwidth}
        \centering
        \includegraphics[width=\textwidth]{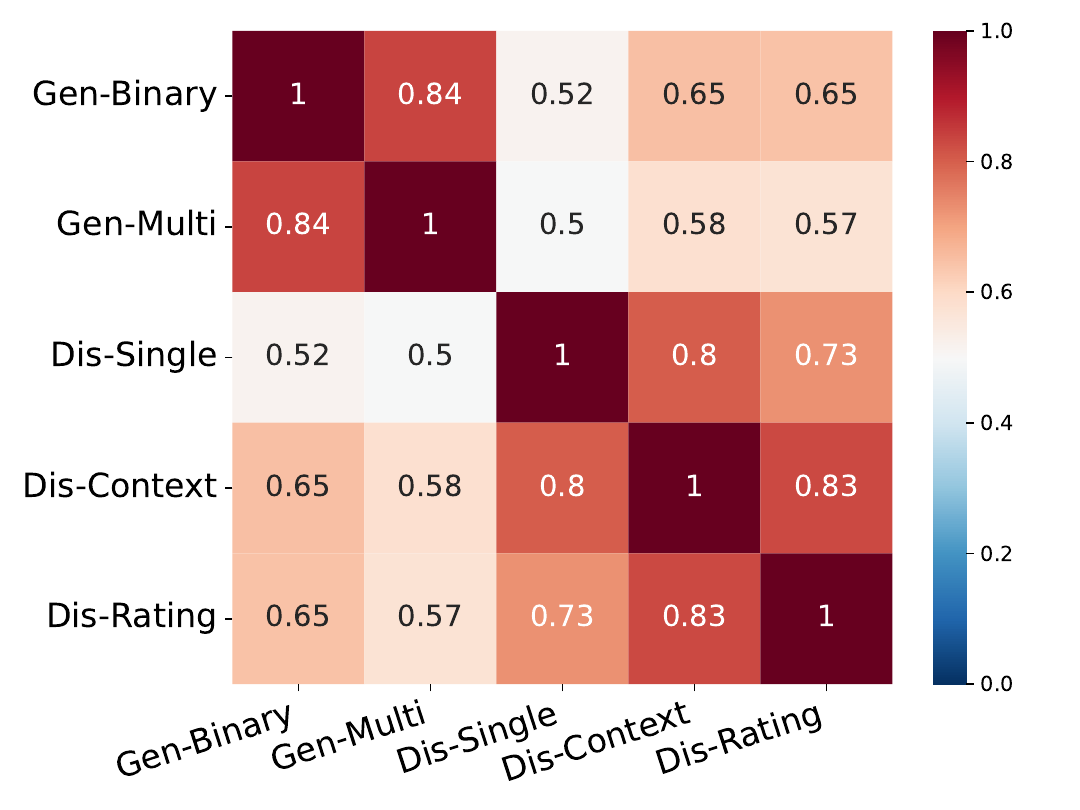}
        \caption{\modelthree\ : response}
        \label{fig:\modelthree_\datasetthree_response}
    \end{subfigure}
    \caption{Heatmaps comparing the Spearman correlation between generative and discriminative confidence elicitation methods for \textit{\datasetthree}. Results shown for \modelone, \modeltwo, and \modelthree.}
    \label{fig:all_models_\datasetthree}
\end{figure*}

\clearpage
\section{Case Study}
\begin{figure*}[h!]
    \centering
    \includegraphics[width=0.80\textwidth]{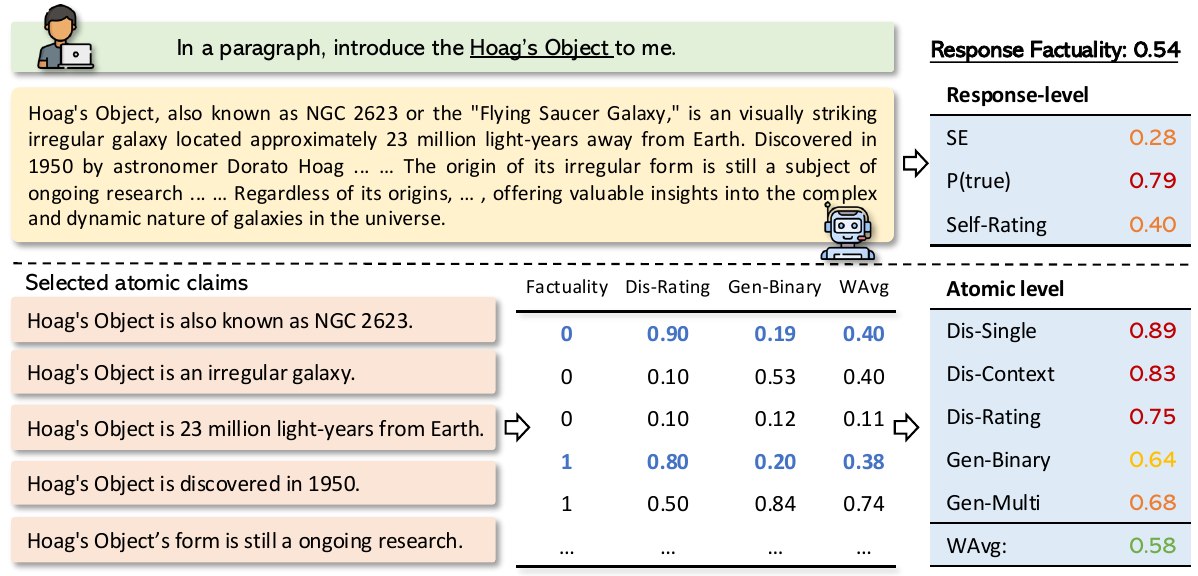}
    \caption{An example from \textit{WildHallu} dataset by Mistral-7B-Instruct. We only select five atomic facts for demonstration. The example shows the effectiveness of calculating confidence in atomic level with fusion strategy.}
    \label{fig:case_study}
\end{figure*}

\section{Prompts} \label{app:prompts}

\begin{table*}[h]
\centering
\begin{tcolorbox}[colback=green!5!white, colframe=green!50!black, title = {Prompts to Generate Responses}]

\textbf{Bios: }

Tell me a bio of <ENTITY>.

    \noindent\begin{tikzpicture}
    \draw[dashed] (0,0) -- (\linewidth,0);
    \end{tikzpicture}

\textbf{LongFact:}

Can you detail the concept of Gravitational Lensing and how it is utilized in modern astronomy for the study and understanding of the universe? \\

Can you explain the process and importance of customer journey mapping in creating effective marketing strategies? 

    \noindent\begin{tikzpicture}
    \draw[dashed] (0,0) -- (\linewidth,0);
    \end{tikzpicture}
    
\textbf{WildHallu:}

In a paragraph, could you tell me what you know about <ENTITY>?

\end{tcolorbox}
\caption{Prompts to generate responses. We use the default prompt template for \textit{Bios} and \textit{WildHallu}. For \textit{LongFact}, it has corresponding Prompt for each question, so we only list two examples here.}
\label{tab:prompt_responses}
\end{table*}

\begin{table*}[h]
\centering
\begin{tcolorbox}[colback=blue!5!white, colframe=blue!50!black, title = {\textsc{Gen-Binary}}]
\ttfamily
Sample: \{sample\}

Sentence: \{sentence\}

Is the sentence supported by the sample above?

You should answer the question purely based on the given sample and not your own knowledge. 

Do not output the explanations.

Your answer should be within ``yes'' or ``no''.

Answer: 

\end{tcolorbox}
\caption{Prompt for \textsc{Gen-Binary}.}
\label{tab:gen_binary}
\end{table*}

\begin{table*}[h!]
\centering
\begin{tcolorbox}[colback=blue!5!white, colframe=blue!50!black, title = {\textsc{Gen-Multi}}]

\ttfamily
Sample: \{sample\}

Sentence: \{sentence\}

Is the sentence supported, refuted, or not mentioned by the sample above?

You should answer the question purely based on the given sample and not your own knowledge.

Do not output the explanations.

Your answer should be within ``supported'', ``refuted'', or ``not mentioned''.

Answer: 

\end{tcolorbox}
\caption{Prompt for \textsc{Gen-Multi}.}
\label{tab:gen_multi}
\end{table*}

\begin{table*}[h!]
\centering
\begin{tcolorbox}[colback=blue!5!white, colframe=blue!50!black, title = {\textsc{Dis-Single}}]
\ttfamily

Please evaluate the truthfulness of the following statement based on your internal knowledge. 

Respond only with ``True'' or ``False'' with no other information.

Statement: \{atomic\_fact\_piece\}

True or False:

\end{tcolorbox}
\caption{Prompt for \textsc{Dis-Single}.}
\label{tab:dis_single}
\end{table*}

\begin{table*}[h!]
\centering
\begin{tcolorbox}[colback=blue!5!white, colframe=blue!50!black, title = {\textsc{Dis-Context}}]
\ttfamily

You are given a topic and a statement related to that topic. 

Evaluate the truthfulness of the statement based solely on your internal knowledge. 

A statement is true only if it is factually accurate according to what you know.

Do not use the topic to evaluate the statement. 

Respond with ``True'' or ``False'' only, without any additional information.

Topic: \{context\}

Statement: \{atomic\_fact\_piece\}

True or False: 

\end{tcolorbox}
\caption{Prompt for \textsc{Dis-Context}.}
\label{tab:dis_context}
\end{table*}

\begin{table*}[h!]
\centering
\begin{tcolorbox}[colback=blue!5!white, colframe=blue!50!black, title = {\textsc{Dis-Rating}}]

\ttfamily

Please rate your confidence in the truth of the following statement on a scale from 0 to 10.

A confidence of 0 means you are not confident at all whether the statement is true, while 10 means you are absolutely confident the statement is true.

The topic provides context for the statement, but focus on the statement itself when rating your confidence.

Provide only the numerical value for your confidence level, bounded by dollar signs (\$), without any additional information.

Topic: \{context\}

Statement: \{atomic\_fact\_piece\}

Confidence: 

\end{tcolorbox}
\caption{Prompt for \textsc{Dis-Rating}.}
\label{tab:dis_rating}
\end{table*}

\end{document}